\documentclass[journal]{IEEEtran}
\usepackage{times}
\usepackage{epsfig}
\usepackage{helvet}
\usepackage{courier}
\usepackage{graphicx}
\usepackage{amsmath}
\usepackage{amssymb}
\usepackage{amsthm}
\usepackage{multirow}
\usepackage{cite}
\usepackage{array}
\usepackage{url}
\usepackage{booktabs}
\usepackage{mathtools}
\usepackage{makecell}
\usepackage{booktabs}
\usepackage{xcolor}

\ifCLASSINFOpdf
\else
\fi

\begin{document}

\indent

© 2022 IEEE. Personal use of this material is permitted. Permission from IEEE must be obtained for all other uses, in any current or future media, including reprinting/republishing this material for advertising or promotional purposes, creating new collective works, for resale or redistribution to servers or lists, or reuse of any copyrighted component of this work in other works. Digital Object Identifier 10.1109/TMM.2022.3164785

%
\title{ChildPredictor: A Child Face Prediction Framework with Disentangled Learning}
%
%
%

\author{Yuzhi~Zhao,~\IEEEmembership{Student~Member,~IEEE,}
        Lai-Man~Po,~\IEEEmembership{Senior~Member,~IEEE,}
        Xuehui~Wang, Qiong~Yan, Wei~Shen, Yujia~Zhang, Wei~Liu, Chun-Kit~Wong, Chiu-Sing~Pang, Weifeng~Ou, Wing-Yin~Yu, Buhua Liu
\thanks{Manuscript received December 6, 2021; Revised February 24, 2022 and March 29, 2022; accepted March 31, 2022; date of current version March 31, 2022. This article was recommended by Associate Editor Chang Xu. \textit{(Corresponding author: Yuzhi Zhao.)}}
\thanks{Y. Zhao, L.-M. Po, Y. Zhang, C.-K. Wong, C.-S. Pang, W. Ou, W.-Y. Yu are with the Department of Electrical Engineering, City University of Hong Kong, Hong Kong, China (e-mail: yzzhao2-c@my.cityu.edu.hk; eelmpo@cityu.edu.hk; yzhang2383-c@my.cityu.edu.hk; ckwong535-c@my.cityu.edu.hk; chiuspang2-c@my.cityu.edu.hk; weifengou2-c@my.cityu.edu.hk; wingyinyu8-c@my.cityu.edu.hk).}
\thanks{X. Wang, W. Shen are with the Artificial Intelligence Institute, Shanghai Jiao Tong University, Shanghai, China (e-mail: wangxuehui@sjtu.edu.cn; wei.shen@sjtu.edu.cn).}
\thanks{Q. Yan is with the SenseTime Research and Tetras.AI, Hong Kong, China. (e-mail: sophie.yanqiong@gmail.com).}
\thanks{W. Liu is with the ByteDance Ltd., Beijing, China (e-mail: liujikun63@gmail.com)}
\thanks{B. Liu is with the Department of Computer Science, Hong Kong Baptist University, Hong Kong, China (e-mail: csbhliu@comp.hkbu.edu.hk)}
\thanks{This article has supplementary material provided by the authors and color versions of one or more figures available at https://doi.org/10.1109/TMM.2022.3164785.}
\thanks{Digital Object Identifier 10.1109/TMM.2022.3164785}
}

\markboth{IEEE Transactions on Multimedia}%
{Shell \MakeLowercase{\textit{Zhao et al.}}: ChildPredictor: A Child Face Prediction Framework with Disentangled Learning}
%

\maketitle

\begin{abstract}

The appearances of children are inherited from their parents, which makes it feasible to predict them. Predicting realistic children’s faces may help settle many social problems, such as age-invariant face recognition, kinship verification, and missing child identification. It can be regarded as an image-to-image translation task. Existing approaches usually assume domain information in the image-to-image translation can be interpreted by ``style'', i.e., the separation of image content and style. However, such separation is improper for the child face prediction, because the facial contours between children and parents are not the same. To address this issue, we propose a new disentangled learning strategy for children’s face prediction. We assume that children’s faces are determined by genetic factors (compact family features, e.g., face contour), external factors (facial attributes irrelevant to prediction, such as moustaches and glasses), and variety factors (individual properties for each child). On this basis, we formulate predictions as a mapping from parents' genetic factors to children's genetic factors, and disentangle them from external and variety factors. In order to obtain accurate genetic factors and perform the mapping, we propose a ChildPredictor framework. It transfers human faces to genetic factors by encoders and back by generators. Then, it learns the relationship between the genetic factors of parents and children through a mapping function. To ensure the generated faces are realistic, we collect a large Family Face Database to train ChildPredictor and evaluate it on the FF-Database validation set. Experimental results demonstrate that ChildPredictor is superior to other well-known image-to-image translation methods in predicting realistic and diverse child faces. Implementation codes can be found at https://github.com/zhaoyuzhi/ChildPredictor.

\end{abstract}


%
\IEEEpeerreviewmaketitle

\section{Introduction}

\IEEEPARstart{T}{HE} appearances of children are inherited from parents. Their internal relations (e.g., kinship verification and identification) have been well studied \cite{xia2011kinship, lu2013neighborhood, dibeklioglu2013like, sun2018video, li2021meta}. It provides the prerequisite for predicting child faces from their parents. Predicting realistic child faces are beneficial to many social issues such as law enforcement, criminal investigations, age-invariant face recognition \cite{park2010age, huang2021age, yan2022age, zhao2022towards}, kinship verification \cite{xia2011kinship, lu2013neighborhood, dibeklioglu2013like, sun2018video, li2021meta}, kinship identification \cite{wang2020kinship}, and missing children identification \cite{chandran2018missing}, especially under the circumstances that \emph{only parent faces are known}. Recently, the generative adversarial network (GAN) \cite{goodfellow2014generative} has shown its advance in the face generation area. If we treat child face prediction as an image- to-image translation issue, there is a potential for GAN to predict high-quality child faces. In this paper, we propose a GAN-based ChildPredictor framework for realistic child face prediction.

\begin{figure*}[t]
\centering
\includegraphics[width=\linewidth]{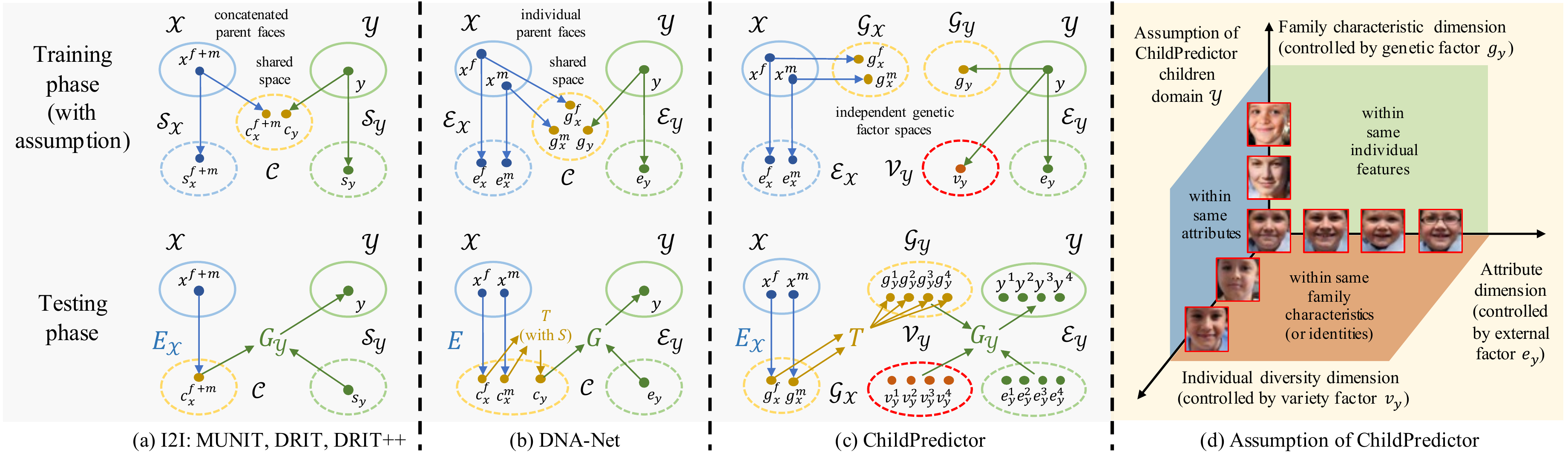}

\vspace{-3mm}

\caption{Assumption and data flow: (a) I2I methods \cite{lee2018diverse, huang2018multimodal, lee2020drit++}; (b) DNA-Net \cite{gao2021dna}; (c), (d) ChildPredictor. The $\mathcal{X}$ and $\mathcal{Y}$ represent parent and children domain, respectively. The $\mathcal{C}$, $\mathcal{S}$, $\mathcal{E}$, $\mathcal{V}$, $\mathcal{G}$ are content, style, external, variety (or attribute for (b)) and genetic latent domains, respectively. The network $E$ and $G$ are pair of encoder and generator that connect image space and latent space. The network $T$ is a mapping function used in the latent space.}
\label{com_arch}

\vspace{-3mm}

\end{figure*}

The fundamental difficulty of child face prediction lies in the requirement on both \emph{diversity} and \emph{similarity} at the same time, in addition to \emph{image quality}. Conditioned on parent faces, the predicted child faces need to be similar to real faces while retaining diversity. The existing strategies to address the issues fall into the two categories:

\begin{enumerate}\setlength{\itemsep}{-0.0cm}
\item Image-to-image translation (I2I) (Figure \ref{com_arch} (a)): Assuming parents' and children's faces form two individual domains, and predicting child faces by transferring ``style'' from parent domain to children domain;

\item DNA-Net \cite{gao2021dna} (Figure \ref{com_arch} (b)): Learning the direct mapping between parents' and children's features, which are generated by the same pre-trained encoder. Diverse prediction is implemented by random selection $S$.
\end{enumerate}

Unfortunately, these methods have difficulties in predicting pleasant faces. Firstly, state-of-the-art I2I methods \cite{lee2018diverse, huang2018multimodal, lee2020drit++} proposed a shared content space and individual style spaces to improve image translation quality. If simply applying it to child face prediction, though the paired parent-child data is used for training I2I methods, they easily fall into ``appearance collapse'' (e.g., the generated children have the same facial structures and contours with parents). We assume that the disentanglement of content (i.e., structure or shape) and style (i.e., texture) is not reasonable for this task since different children could have similar structures with parent faces but not the same. Secondly, DNA-Net fused features of parents and then changed them using an age regression model \cite{zhang2017age}. Since parent and child faces share the same latent space, it restricts the feature representation and reality. Though it applied a random selection $S$ to combine mothers' and fathers' features, the framework is trained in a unimodal fashion. Therefore, the results are not diverse enough.

This paper presents a novel framework for synthesizing realistic and diverse child faces. Built upon previous experiences, we first assume a human face is determined by a \emph{genetic factor, external factor, and variety factor}, among which a child's genetic factor is predicted from parents' genetic factors. These three factors are trained to be \emph{orthogonal}, as shown in Figure \ref{com_arch} (d), and detailed definitions are concluded as:

\begin{enumerate}\setlength{\itemsep}{-0.0cm}
\item Genetic factor $g$: Family characteristics such as contour, eye shape, pupil color, skin color, etc., which model the inter-identity variation between different identities;

\item External factor $e$: Categorical and gene-irrelevant attributes, e.g., moustaches, ages, expressions, glasses;

\item Variety factor $v$: Individual diversities for an identity. It is not correlated to genetic and external factors, which models acquired factors (intra-identity variations).
\end{enumerate}

Based on the assumption, we propose a GAN-based framework called ChildPredictor to predict child faces from their parents. The data flow of ChildPredictor is shown in Figure \ref{com_arch} (c). There are two sequential steps in the training phase compared with previous one-step pipelines:

\begin{enumerate}\setlength{\itemsep}{-0.0cm}
\item Domain-specific disentangled learning: Learning disentangled representations of different factors for both domains, i.e., the disentanglement of $\mathcal{G_X}$/$\mathcal{E_X}$ and $\mathcal{G_Y}$/$\mathcal{E_Y}$/$\mathcal{V_Y}$ (please see the caption of Figure \ref{com_arch} and the training phase of Figure \ref{com_arch} (c) for their definitions), respectively;

\item Inter-domain multimodal mapping: Predicting multiple children's genetic factors between the disentangled genetic factors in both spaces $\mathcal{G_X}$ and $\mathcal{G_Y}$.
\end{enumerate}

There are four advances of our setting: 1) Compared with the general content-style separation \cite{huang2018multimodal}, the proposed genetic-external-variety disentangled learning, which is a three-way decomposition specially designed for the task, considers inter-identity variation, gene-irrelevant attributes, and intra-identity variation; 2) Since the data distribution of parents' and children's faces are quite different, training in each domain can lead to better domain-specific features than joint training \cite{gao2021dna}; 3) Since domain-specific training is not based on limited parent-child pairs, additional data can be used, e.g., FFHQ child faces \cite{karras2019style} are used to improve the generation quality of $\mathcal{G_Y}$; 4) Our mapping function learns a multimodal prediction instead of unimodal \cite{gao2021dna} thus it is able to simulate different child identities from the same parent. In summary, our ChildPredictor framework largely improves child face prediction quality in similarity, diversity, and realness.


In addition, to train ChildPredictor, we newly collect a large-scale FF-Database from the Internet. It includes 7488 parent faces and 8558 child faces with 128$\times$128 resolution and even gender distribution. Each face is aligned to be almost frontal and labeled with 6 attributes which are gender, age, expression, glasses, moustache, and skin color. For evaluation, we notice that common metrics (e.g., PSNR) fail to measure face similarity since the generated and ground truth faces are not pixel-wisely aligned. We use a cosine similarity metric to compute feature distances from a pre-trained face recognition model and conduct a human perceptual study for subjective evaluation. The experiment results on the FF-Database validation set demonstrate that ChildPredictor performs better than previous pipelines \cite{zhu2017unpaired, yi2017dualgan, liu2017unsupervised, lee2018diverse, huang2018multimodal, lee2020drit++, gao2021dna} on cosine similarity, FID \cite{Heusel2017GANs}, LPIPS \cite{Zhang2018The} scores and human preference rates.

The main contributions of this paper are as follows:

\begin{enumerate}\setlength{\itemsep}{-0cm}
\item We propose a GAN-based ChildPredictor framework specializing in the task of child face prediction;

\item We propose a latent representation disentangled learning method by sampling from three latent factors;

\item We newly collect a large-scale Family Face Database (FF-Database). To our best knowledge, it is the first dataset for child face prediction with labeled attributes;

\item We propose a cosine similarity metric and a human perceptual study for evaluating paired predicted and real faces. The ChildPredictor achieves the best performance.
\end{enumerate}

\section{Related Work}

\textbf{Generative Adversarial Network (GAN) and Inversion.} The GAN \cite{goodfellow2014generative} has accomplished great improvements in image generation. It consists of a generator and a discriminator, where the generator produces realistic samples and the discriminator distinguishes input samples are from ground truth or generated. However, GAN is hard to converge and unstable. To address the issue, some methods minimized mode collapse \cite{mao2019mode} and loss fluctuation \cite{arjovsky2017wasserstein}, or proposed new architectures \cite{mirza2014conditional, karras2018progressive, karras2019style, karras2020analyzing}. To discover the relation between generated images and latent space, GAN inversion techniques have been widely studied. For instance, \cite{Goetschalckx2019GANalyze} and \cite{Jahanian2019On} learned to search interpretable directions in GAN’s latent space. To edit images accurately, inverse encoders \cite{bau2019inverting, gabbay2019style, abdal2019image2stylegan, abdal2020image2stylegan++, shen2020interpreting, viazovetskyi2020stylegan2, richardson2021encoding} were proposed to simulate reverse process of GANs. They learned the inversion in original latent space or extend W+ space.

\textbf{Face Attribute Transfer.} The previous face attribute transfer methods are normally based on paired data \cite{langner2010presentation}. However, it is hard to collect many different attributes for the same person. To avoid that, researchers used several representations such as pre-defined attributes \cite{Perarnau2016, zhou2017genegan, xiao2018dna, zhang2018generative, choi2018stargan}, facial landmarks \cite{song2018geometry}, face mask \cite{li2018beautygan}, and action unit \cite{pumarola2018ganimation}. These methods were further enhanced by processing multiple attributes simultaneously \cite{xiao2018elegant, he2019attgan, he2020pa}. More recently, the GAN inversion techniques \cite{nitzan2020face, tov2021designing} and latent disentangled learning \cite{zhu2020learning, zhu2021and, he2021eigengan} have been used in the face attribute transfer area. For instance, Nitzan et al. \cite{nitzan2020face} transferred attributes based on an exemplar face by manipulating pre-trained GAN's latent space.


\textbf{Image-to-Image Translation (I2I) and Disentangled Learning.} The I2I denotes the mappings of images from one domain to other domains. For instance, Pix2Pix \cite{isola2017image} used a conditional GAN to perform domain transfer on pixel-aligned data, which is necessary, otherwise, the results are blurry. To extend I2I to unaligned data, some approaches have been developed, e.g., enlarging the distance between generated samples and source  \cite{shrivastava2017learning, bousmalis2017unsupervised} and cycle consistency \cite{zhu2017unpaired, kim2017learning, yi2017dualgan}. However, the models often fail (i.e., mode collapse) when there exist extreme transformations or training data is limited.

To improve the image translation quality, the disentangled learning \cite{liu2017unsupervised, lee2018diverse, huang2018multimodal, choi2020stargan, lee2020drit++, lin2020tuigan, chen2020distilling, zhuang2020dunit, liu2021smoothing, fabio2021CoMoGAN, li2021hisd, emami2020spa, kwong2021unsupervised} assumed that images from individual domains share the same latent content space but separate style space. By changing ``style code'', the network transfers input images to other domains while maintaining the content information. To further address the mode collapse issue, \cite{mao2019mode} proposed a \emph{mode-seeking loss} to enlarge distances of different generated samples for regularization. However, such disentangled learning methods are not appropriate for child face prediction since the goal is not only to transfer the style but also to consider facial attributes and intra-identity variations. To address the issue, we propose a new disentangled learning method based on genetic, external, and variety factors (please see Figure \ref{com_arch} (c) and (d) for their definitions).

\section{Data Collection of FF-Database}

\begin{figure}[t]
\centering
\includegraphics[width=\linewidth]{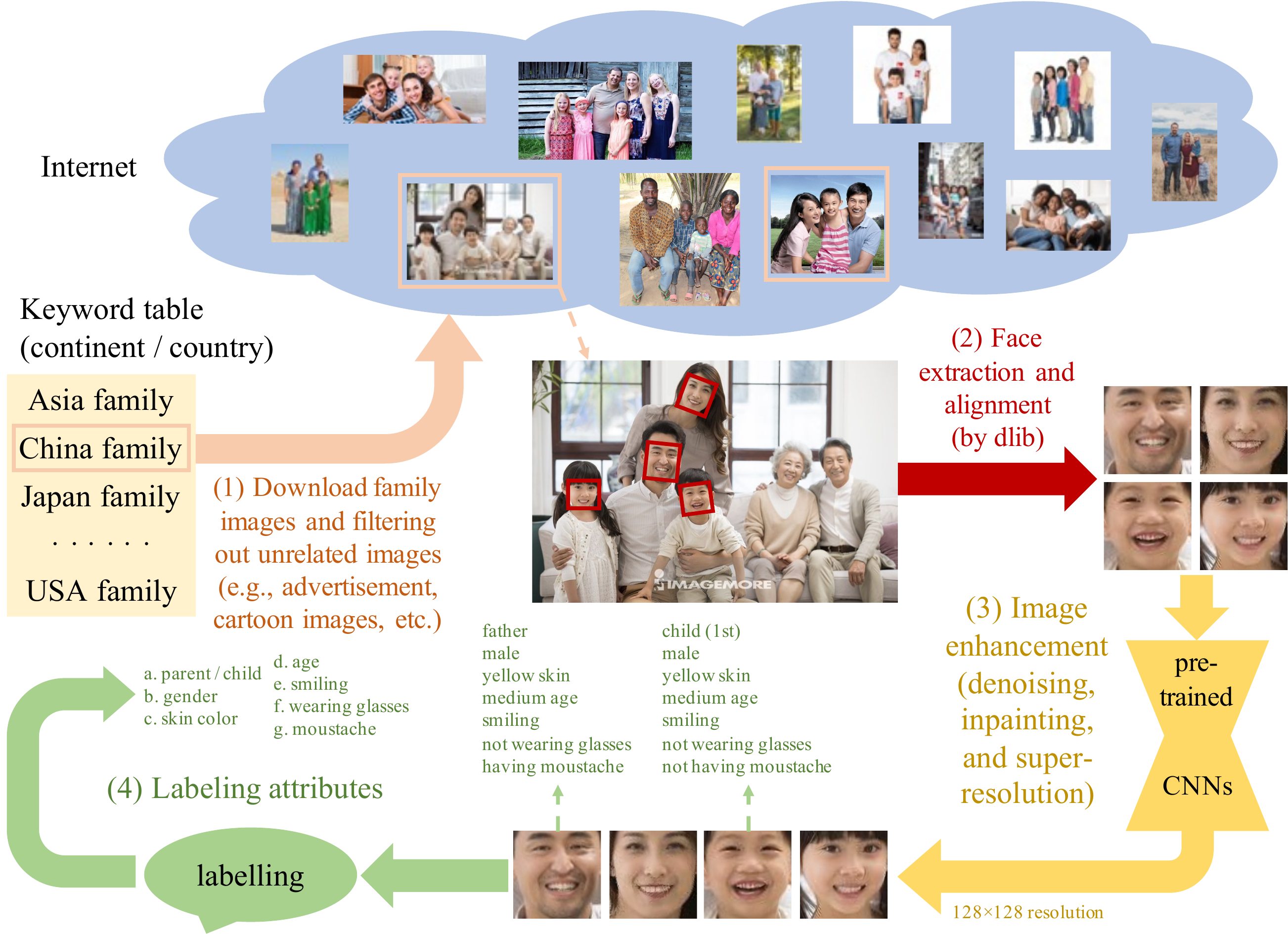}

\vspace{-2mm}

\caption{Illustration of image collection workflow for the FF-Database.}
\label{dataset}

\vspace{-2mm}

\end{figure}

We collect a large-scale Family Face Database (FF-Database), consisting of 16046 images with 128$\times$128 resolution. Built upon it, we learn the child prediction in a data-driven manner. There are 4 steps to collect a group of images in FF-Database, as shown in Figure \ref{dataset}: 1) Downloading family images by country or district names in 6 continents from the Internet, and filtering out unrelated face images; 2) Extracting faces by dlib$\footnote{http://dlib.net/}$ and aligning them to be almost frontal; 3) Enhancing faces by denoising \cite{gu2019self}, inpainting \cite{yu2019free}, super-resolution \cite{wang2018esrgan} algorithms, and resizing to 128$\times$128 resolution. Note that the very low-quality images are discarded by human effort; 4) Labeling them with 6 attributes including gender, age, expression, glasses, moustache, and skin color.

We divide the whole dataset into two parts, where the training set includes 15538 faces and the validation set includes 708 faces. Specifically, there are 7148 parents and 8190 children in the training set; and there are 340 and 368 faces in the validation set, respectively. The attributes and the division of training and validation sets are concluded in Table \ref{ffdatabase}, respectively.

\begin{table}[t]
\caption{Conclusion of the labeled facial attributes of the FF-Database training and validation sets.}

\vspace{-3mm}

\label{ffdatabase}
\begin{center}
\begin{tabular}{lccccccc}
\hline
\multirow{2}{*}{Tag} & \multirow{2}{*}{Attribute} & \multicolumn{3}{c}{Parent faces} & \multicolumn{3}{c}{Child faces} \cr & & True & False & Total & True & False & Total \\
\hline
Train & Male & 3574 & 3574 & 7148 & 4104 & 4086 & 8190 \\
Train & Young & 804 & 6344 & 7148 & 1043 & 7147 & 8190 \\
Train & Smile & 6384 & 764 & 7148 & 5860 & 2330 & 8190 \\
Train & Glass & 572 & 6576 & 7148 & 279 & 7911 & 8190 \\
Train & Moustache & 1281 & 5867 & 7148 & 54 & 8136 & 8190 \\
\hline
Val & Male & 170 & 170 & 340 & 192 & 176 & 368 \\
Val & Young & 8 & 332 & 340 & 35 & 333 & 368 \\
Val & Smile & 334 & 6 & 340 & 352 & 16 & 368 \\
Val & Glass & 12 & 328 & 340 & 6 & 362 & 368 \\
Val & Moustache & 53 & 287 & 340 & 0 & 368 & 368 \\
\hline

\vspace{-6mm}

\end{tabular}
\end{center}
\end{table}

\begin{figure*}[t]
\centering
\includegraphics[width=\linewidth]{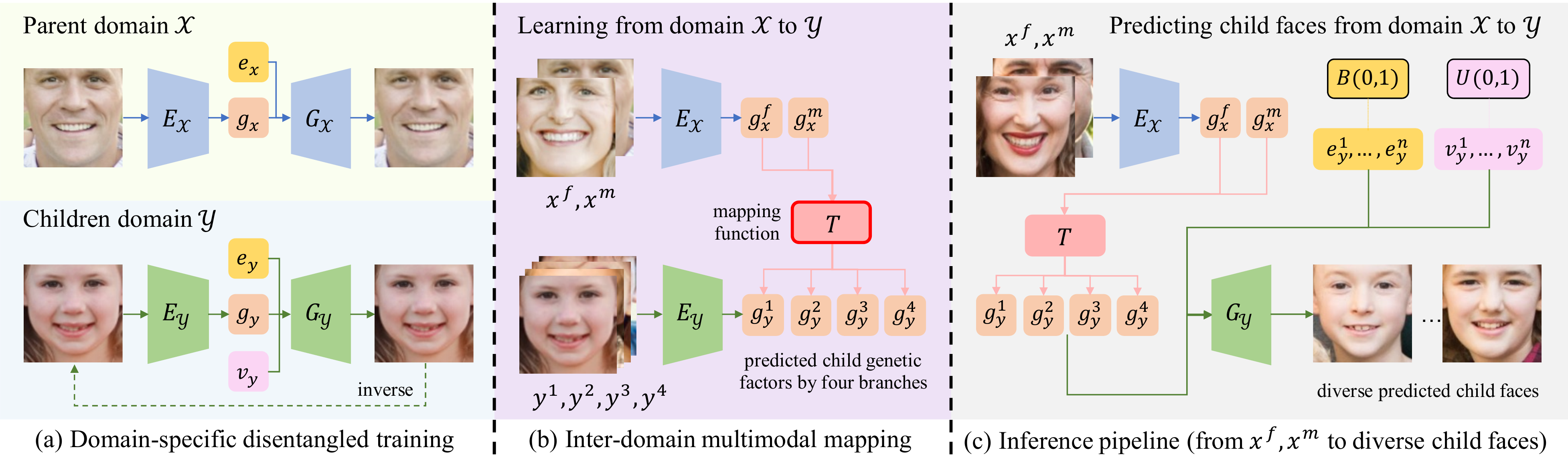}

\vspace{-2mm}

\caption{Illustration of (a) self-supervised disentangled learning (1st training step); (b) inter-domain multimodal mapping (2nd training step); (c) inference pipeline. The latent codes $g_*$, $e_*$, $v_\mathcal{Y}$ represent genetic, external and variety factors.}
\label{generator}

\vspace{-2mm}

\end{figure*}

\section{Methodology}

\begin{table}[t]
\caption{List of all the notations used in the paper.}

\vspace{-4mm}

\label{notation}
\begin{center}
\begin{tabular}{ll}
\hline
Notation & Definition \\
\hline
$D_\mathcal{X}$ & Discriminator for parent domain $\mathcal{X}$ \\
$D_\mathcal{G_X}$ & Discriminator for parent genetic factor distribution $\mathcal{G_X}$ \\
$D_\mathcal{Y}$ & Discriminator for children domain $\mathcal{Y}$ \\
$e^f_x$ & External factor of father face image $x^f$ \\
$e^m_x$ & External factor of mother face image $x^m$ \\
$\mathcal{E_X}$ & Distribution of parent external factor $e^f_x$ and $e^m_x$ \\
$e_y$ & External factor of child face image $y$ \\
$\mathcal{E_Y}$ & Distribution of child external factor $e_y$ \\
$E_\mathcal{X}$ & Encoder from $\mathcal{X} \rightarrow \mathcal{G_X}$ of ChildPredictor \\
$E_\mathcal{Y}$ & Encoder from $\mathcal{Y} \rightarrow \mathcal{G_Y}$ of ChildPredictor \\
$g^f_x$ & Genetic factor of father face image $x^f$ \\
$g^m_x$ & Genetic factor of mother face image $x^m$ \\
$\mathcal{G_X}$ & Distribution of parent genetic factor $g^f_x$ and $g^m_x$ \\
$g_y$ & Genetic factor of child face image $y$ \\
$\mathcal{G_Y}$ & Distribution of child genetic factor $g_y$ \\
$G_\mathcal{X}$ & Generator from $\mathcal{G_X} \rightarrow \mathcal{X}$ of ChildPredictor \\
$G_\mathcal{Y}$ & Generator from $\mathcal{G_Y} \rightarrow \mathcal{Y}$ of ChildPredictor \\
$L_*$ & Loss functions for training the ChildPredictor \\
$\lambda_*$ & Trade-off parameters for different loss functions \\
$T$ & Mapping function from $\mathcal{G_X} \rightarrow \mathcal{G_Y}$ of ChildPredictor \\
$v_y$ & Variety factor of child face image $y$ \\
$\mathcal{V_Y}$ & Distribution of child variety factor $g_y$ \\
$x^f$ & Father face image \\
$x^m$ & Mother face image \\
$\widehat{x_{e}}$ & Generated parent face image according to $e_x$ \\
$\mathcal{X}$ & Parent domain (the set of parent face images) \\
$y$ & Child face image \\
$\widehat{y_{e,v}}$ & Generated child face image according to $e_y$ and $v_y$ \\
$\mathcal{Y}$ & Children domain (the set of child face images) \\
\hline

\vspace{-8mm}

\end{tabular}
\end{center}
\end{table}

\subsection{Problem Formulation}

Given paired parent faces $x^f,x^m \in \mathcal{X}$ and a child face $y \in \mathcal{Y}$, the target of child face prediction is to learn $p(y|x^f,x^m)$. Note that their corresponding genetic factors $g_y, g^f_x, g^m_x$ are the compact and simplified representations of face images by our definition. Therefore, it is more tractable to solve the problem in the genetic domain, i.e., to learn $p(g_y|g^f_x, g^m_x)$, only if the genetic factors and faces are transferable from each other. To achieve this, we design the two-stage framework, \textbf{domain-specific disentangled learning} which learns to extract the genetic factors from faces and restore them to faces for parent and children domain separately, and \textbf{inter-domain multimodel mapping} which maps parent genetic factors to children domain.

Our framework is different from existing methods, as shown in Figure \ref{com_arch}. Firstly, the I2I methods assume a shared content space for parent and children domains and easily falls into appearance collapse. However, the child face prediction is not simply transferring styles of the faces. Secondly, DNA-Net assumes a shared latent space for parent and child faces and then performs age-regression. However, this design cannot ensure the generator recovers realistic child faces. As for the ChildPredictor, we assume parent and child genetic factors are in the individual latent spaces and propose a mapping function $T$ to learn the prediction from parents' genetic factors to child genetic factors.

During inference, our ChildPredictor firstly extracts genetic factors from the parent faces, then maps them to children genetic domain, finally predicts diverse outputs by sampling different genetic, external, and variety factors. The workflow is shown in Figure \ref{generator} (c) and all the notations are concluded in Table \ref{notation}. Details on the training process are presented below.


\subsection{Domain-specific Disentangled Learning}

\subsubsection{Assumption}

We perform the domain-specific disentangled learning in the parent and children domain separately. To ensure the transferability between face images and genetic factors for both domains, the two encoder-generator pairs ($E_\mathcal{X}$, $G_\mathcal{X}$), ($E_\mathcal{Y}$, $G_\mathcal{Y}$) must satisfying:
\begin{equation}
x^f = G_\mathcal{X} ( E_\mathcal{X} (x^f), e_x^f ), E_\mathcal{X} (x^f) = g_x^f,
\label{311_1}
\end{equation}
\begin{equation}
x^m = G_\mathcal{X} ( E_\mathcal{X} (x^m), e_x^m ), E_\mathcal{X} (x^m) = g_x^m,
\label{311_2}
\end{equation}
\begin{equation}
y = G_\mathcal{Y} ( E_\mathcal{Y} (y), e_y, v_y ), E_\mathcal{Y} (y) = g_y,
\label{312}
\end{equation}
where $e_*$ and $v_*$ denote external and variety factors, respectively. The goal is to ensure that encoders should only extract genetic factors. To further differentiate their roles, we assume genetic factors follow normal distribution and variety factors follow uniform distribution. External factors are categorical attributes thus binary distributed.


\subsubsection{Parent Domain $\mathcal{X}$}

We adopt \cite{ronneberger2015u} as $E_\mathcal{X}$ and $G_\mathcal{X}$, as shown in Figure \ref{generator} (a). The encoder $E_\mathcal{X}$ firstly encodes a parent image $x$ to a genetic factor $\widehat{g_x}$. Then, the generator $G_\mathcal{X}$ receives the produced genetic factor with a given external factor $e_x$ to recover a face image $\widehat{x_e}$. It is an identity transformation only when the given $e_x$ is from $x$; otherwise, $G_\mathcal{X}$ performs attribute transfer while maintaining identity unchanged. It is because genetic factors are disentangled from external factors. The \emph{whole loss} $L_\mathcal{X}$ for training $E_\mathcal{X}$ and $G_\mathcal{X}$ is given as:
\begin{equation}
L_\mathcal{X} = \lambda_\mathcal{X}^1 L_\mathcal{X}^1 + \lambda_\mathcal{X}^2 L_\mathcal{X}^C + \lambda_\mathcal{X}^3 L_\mathcal{X}^G + \lambda_\mathcal{G_X}^4 L_\mathcal{G_X}^G,
\label{321}
\end{equation}
where $\lambda_\mathcal{X}^1$, $\lambda_\mathcal{X}^2$, $\lambda_\mathcal{X}^3$ and $\lambda_\mathcal{G_X}^4$ are trade-off parameters. The \emph{L1 reconstruction loss $L_\mathcal{X}^1$} maintains attribute-invariant features determined by genetic factors. The \emph{classification loss $L_\mathcal{X}^C$} enlarges the differences between faces with and without attributes determined by different dimensions of external factors, which share similar definitions with \cite{he2019attgan}. The \emph{image-level adversarial loss $L_\mathcal{X}^G$/$L_\mathcal{X}^D$} \cite{arjovsky2017wasserstein} enhances perceptual similarity. To further disentangle genetic and external factors in the latent space, we use a \emph{genetic-factor-level adversarial loss $L_\mathcal{G_X}^G$/$L_\mathcal{G_X}^D$} to make $\mathcal{G_X}$ approach to standard normal distribution. The detailed formulations are:
\begin{equation}
L_\mathcal{X}^1 = \mathbb{E} [|| x - \widehat{x_e} ||_1],
\label{321_detail1}
\end{equation}
\begin{equation}
\begin{aligned}
L_\mathcal{X}^C &= - e_x log ( D_\mathcal{X}^{C}(x) ) + (1 - e_x) log ( 1 - D_\mathcal{X}^{C}(x) ) \\
&- e_x log ( D_\mathcal{X}^{C}(\widehat{x_e}) ) + (1 - e_x) log ( 1 - D_\mathcal{X}^{C}(\widehat{x_e}) ),
\label{321_detail2}
\end{aligned}
\end{equation}
\begin{equation}
\begin{aligned}
&L_\mathcal{X}^G = - \mathbb{E} [ D_\mathcal{X}^{G}(\widehat{x_e}) ], \\
&L_\mathcal{X}^D = \mathbb{E} [ D_\mathcal{X}^{G}(\widehat{x_e}) ] - \mathbb{E} [ D_\mathcal{X}^{G}(x) ],
\label{321_detail3}
\end{aligned}
\end{equation}
\begin{equation}
\begin{aligned}
&L_\mathcal{G_X}^G = - \mathbb{E} [ D_\mathcal{G_X}(\widehat{g_x}) ], \\
&L_\mathcal{G_X}^D = \mathbb{E} [ D_\mathcal{G_X}(\widehat{g_x}) ] - \mathbb{E} [ D_\mathcal{G_X}(u) ],
\label{321_detail4}
\end{aligned}
\end{equation}
where two discriminators are used to compute these losses. Image-level discriminator $D_\mathcal{X}$ has two branches, one of which outputs a binary vector for computing $L_\mathcal{X}^C$ and the other of which outputs a scalar for computing $L_\mathcal{X}^G$. Factor-level discriminator $D_\mathcal{G_X}$ receives genetic factors. In Equation \ref{321_detail4}, $u$ is a random sample from a standard Gaussian distribution $N(0,1)$ with the same dimension of genetic factors $\widehat{g_x}$. Note that the variety factor is not considered in the parent domain as there is only one parent pair for any child, while there is more than one child for each parent pair.

\subsubsection{Children Domain $\mathcal{Y}$}

We adopt PGGAN as $G_\mathcal{Y}$ to achieve a higher-quality generation. It generates images based on the given genetic factors, external factors, and variety factors, as shown in Figure \ref{generator} (a). To perform disentangled learning, we use a GAN-inverse encoder $E_\mathcal{Y}$ to recover only genetic factors from images generated by a pre-trained $G_\mathcal{Y}$.

To pre-train $G_\mathcal{Y}$, we define the \emph{loss} $L_{G_\mathcal{Y}}$ as:
\begin{equation}
L_{G_\mathcal{Y}} = \lambda_\mathcal{Y}^1 L_\mathcal{Y}^G + \lambda_\mathcal{Y}^2 L_\mathcal{Y}^C + \lambda_\mathcal{Y}^3 L_\mathcal{Y}^M,
\label{335}
\end{equation}
where $L_\mathcal{Y}^G$ is an \emph{adversarial loss} \cite{goodfellow2014generative}, which promotes $G_\mathcal{Y}$ generating realistic faces. $L_\mathcal{Y}^C$ is the \emph{auxiliary classification loss} \cite{odena2017conditional}, which ensures external factors can control the facial attributes. $L_\mathcal{Y}^M$ is the \emph{mode-seeking loss} \cite{mao2019mode} which ensures variety factors are related to individual variations. Their detailed definitions are as follows:
\begin{equation}
\begin{aligned}
&L_\mathcal{Y}^G = - \mathbb{E} [ D_\mathcal{Y}^G(\widehat{y_{e,v}}) ], \\
&L_\mathcal{Y}^D = \mathbb{E} [ D_\mathcal{Y}^G(\widehat{y_{e,v}}) ] - \mathbb{E} [ D_\mathcal{Y}^G(y) ],
\end{aligned}
\label{335_detail1}
\end{equation}
\begin{equation}
\begin{aligned}
L_\mathcal{X}^C &= - e_y log ( D_\mathcal{Y}^C(y) ) + (1 - e_y) log ( 1 - D_\mathcal{Y}^C(y) ) \\
&- e_y log ( D_\mathcal{Y}^C(\widehat{y_{e,v}}) ) + (1 - e_y) log ( 1 - D_\mathcal{Y}^C(\widehat{y_{e,v}}) ),
\end{aligned}
\label{335_detail2}
\end{equation}
\begin{equation}
L_\mathcal{Y}^M = \max \limits_{G_\mathcal{Y}} ( \frac{d_\mathcal{Y} ( \widehat{y_{e,v^2}} - \widehat{y_{e,v^1}} ) } { d_\mathcal{V_Y} ( v^2_y - v^1_y ) } ),
\label{335_detail3}
\end{equation}
where $\widehat{y_{e,v}}$ is the randomly generated result based on an external factor $e_y$ and a variety factor $v_y$. $y$ is a real sample selected from all training child images. $D_\mathcal{Y}^G$ and $D_\mathcal{Y}^C$ are two branches of the discriminator $D_\mathcal{Y}$. When computing $L_\mathcal{Y}^M$, there are two individual outputs $\widehat{y_{e,v^1}}$ and $\widehat{y_{e,v^2}}$ generated from the same external factor $e_y$ but individual variety factors $v^1_y$ and $v^2_y$. We adopt the $l_1$-norm as distance metric $d_*(\cdot)$, which includes 3 sequential operations: subtraction, taking absolute value, and computing average.


Then, we train $E_\mathcal{Y}$ to disentangle the genetic factor from the other two factors in the latent space of $G_\mathcal{Y}$. Suppose that a fixed genetic factor is used to generate faces with different external and variety factors, an ideal encoder can restore the same genetic factor (i.e., same identity) from those faces. Based on the assumption, we train the $E_\mathcal{Y}$ by \emph{loss} $L_{E_\mathcal{Y}}$:
\begin{equation}
L_{E_\mathcal{Y}} = \mathbb{E} [ || g_y - E_\mathcal{Y}( G_\mathcal{Y} (g_y, e_y, v_y) ) ||_1 ],
\label{336}
\end{equation}
where the $G_\mathcal{Y}$ is fixed. The $e_y$ and $v_y$ are randomly sampled for a same $g_y$. After its convergence, we can obtain the disentangled genetic factors for arbitrary child faces by $E_\mathcal{Y}$.

\subsection{Inter-domain Multimodal Mapping}

Different from previous I2I methods, we perform the ``multimodal mapping'' from parent domain to children domain only on genetic factors in the latent space. The reason is after the disentangled learning for domain $\mathcal{X}$ and $\mathcal{Y}$, the genetic factors $g_*$ are disentangled from other factors $e_*$ and $v_*$, and by our definition only the genetic factors among the three are related between the two domains.



To learn $\mathcal{G_X} \rightarrow \mathcal{G_Y}$, we firstly obtain parent-child genetic factor pairs ($g_x^f$/$g_x^m$, $g_y$) for each family by encoding faces through $E_\mathcal{X}$ and $E_\mathcal{Y}$. Then, we use a neural network as mapping function $T$, as shown in Figure \ref{generator} (b). To simulate multiple children, we simply let $T$ predict $k$ different genetic factors by $k$ branches, i.e., $T: \widehat{g_y^1}, \widehat{g_y^2},..., \widehat{g_y^k} = T ( g_x^f, g_x^m )$. We set $k=4$ in our experiments. The training for $T$ is supervised by parent-child genetic factor pairs and the following loss $L_T$:
\begin{equation}
\begin{aligned}
L_T &= \mathbb{E} [ || \widehat{g_y^1} - g_y^1 ||_1 ] + \lambda_T^1 \mathbb{E} [ || \widehat{g_y^2} - g_y^2 ||_1 ] \\
&+ \lambda_T^2 \mathbb{E} [ || \widehat{g_y^3} - g_y^3 ||_1 ] + \lambda_T^3 \mathbb{E} [ || \widehat{g_y^4} - g_y^4 ||_1 ],
\end{aligned}
\label{341}
\end{equation}
where $\widehat{g_y^j}$ and ${g_y^j}$ are the $j$-th predicted and real genetic factor, respectively. Note that different loss coefficients (1, $\lambda_T^1$, $\lambda_T^2$, and $\lambda_T^3$) are used for the 4 predictions in order to fulfill the multimodal prediction. For families with 4 children or more, we select the first 4 children as ground truth $g_y^1$, $g_y^2$, $g_y^3$, and $g_y^4$. For families with less than 4 children, we replace the blank position $g_y^j$ with $g_y^1$. For instance, for families with only 3 children, we use the 1-st child as ground truth for the 4-th branch. Note that, different loss coefficients are applied to 1-st and 4-th branches though their ground truth images are the same; while the ground truth images of 2-nd and 3-rd branches are different from 1-st and 4-th branches. Therefore, different optimizations are achieved for all 4 branches.



\begin{table*}[t]
\caption{The details for the training steps. The ``Network'', ``LR'', ``Epoch'' and ``Loss'' represent networks trained in this step, current learning rate, total epochs and losses used in the step, respectively. The ``All'' denotes all networks or all losses are used.}
\label{training_step_details}

\vspace{-5mm}

\begin{center}
\begin{tabular}{c|ccccc}
\hline
Step & Network & LR & Epoch & Loss & Dataset \\
\hline
1 & $E_\mathcal{X}$, $G_\mathcal{X}$ & $2\times10^{-4}$ & 200 & $L_\mathcal{X}^1$, $L_\mathcal{X}^C$, $L_\mathcal{X}^G$ & FF-Database parent images \\
1 & $G_\mathcal{Y}$ & $1.5\times10^{-3}$ & 200 & $L_\mathcal{Y}$ ($L_\mathcal{Y}^G$, $L_\mathcal{Y}^C$, $L_\mathcal{Y}^M$) & FF-Database and FFHQ child images \\
\hline
2 & $E_\mathcal{X}$, $G_\mathcal{X}$ & $2\times10^{-4}$ & 200 & $L_\mathcal{X}$ ($L_\mathcal{X}^1$, $L_\mathcal{X}^C$, $L_\mathcal{X}^G$, $L_\mathcal{Z_X}^G$) & FF-Database parent images \\
2 & $E_\mathcal{Y}$ & $1\times10^{-4}$ & 200 & $L_{E_\mathcal{Y}}$ & no data needed ($E_\mathcal{Y}$ is trained by sampling from latent codes) \\
\hline
3 & $T$ & $1\times10^{-3}$ & 200 & $L_T$ & FF-Database parents-children image pairs \\
\hline
4 & All & $1\times10^{-5}$ & 10 & All & FF-Database parents-children image pairs \\
\hline
\end{tabular}
\end{center}

\vspace{-6mm}

\end{table*}

\subsection{Network Architecture}

\subsubsection{Parent Domain $\mathcal{X}$}

The parent domain networks consist of encoder-decoder pair $E_\mathcal{X}$, $G_\mathcal{X}$ and two discriminators $D_\mathcal{X}$/$D_\mathcal{G_X}$. $E_\mathcal{X}$ has 5 generator convolutional blocks (including a convolutional layer, a BatchNorm \cite{ioffe2015batch} layer, and a LeakyReLU \cite{maas2013rectifier} activation function) to produce genetic factors $g_x$. $G_\mathcal{X}$ adopts 5 generator convolutional blocks to progressively upsample the combinations of genetic and external factors. The feature maps of the encoder are injected into the generator as U-Net \cite{ronneberger2015u}. $D_\mathcal{X}$ has 5 discriminator convolutional blocks (including a transposed convolutional layer, a InstanceNorm \cite{ulyanov2016instance} layer, and a ReLU \cite{nair2010rectified} activation function), followed by 2 parallel MLP layers. The outputs are used for computing \emph{WGAN adversarial loss $L_\mathcal{X}^G$} and \emph{classification loss $L_\mathcal{X}^G$}, respectively. $D_\mathcal{G_X}$ contains 2 discriminator convolutional blocks followed by a MLP layer. The output is adopted to compute the \emph{genetic factor adversarial loss $L_\mathcal{G_X}$}.

\subsubsection{Children Domain $\mathcal{Y}$}

We modify the official PGGAN \cite{karras2018progressive} as $G_\mathcal{Y}$, which receives genetic, external, and variety factors as inputs. The factors are concatenated along the channel dimension. The output resolution is 128$\times$128. We use PixelNorm \cite{karras2018progressive} and PReLU \cite{he2015delving} as the normalization layer and activation function, respectively. $E_\mathcal{Y}$ inverses an image generated by $G_\mathcal{Y}$ back to the genetic factor. It is composed of a VGG-network \cite{simonyan2014very} and a MLP layer, where the MLP layer projects the features from VGG-network into 480-dimensional output, which is consistent to the dimension of genetic factor.

\subsubsection{Mapping Function $T$}

The mapping network $T$ receives the genetic factors of mother and father as the inputs. It consists of a head module, a body module, and a tail module. The head module is a simple combination of two convolutional layers and a LeakyReLU activation function. The body module contains 5 residual layers \cite{he2016deep} with LeakyReLU activation function. The tail module includes 4 MLP layers which predicts 4 genetic factors ($\widehat{g_y^1}$, $\widehat{g_y^2}$, $\widehat{g_y^3}$, and $\widehat{g_y^4}$). In addition, we apply a normalization operation (minus mean and divide variance for each output genetic factor) to the output genetic factors to push them to standard Guassian distribution.

%

\section{Experiment}

\subsection{Training Details}

\textbf{General Training Details.} The training process of ChildPredictor can be concluded in 4 steps, as shown in Table \ref{training_step_details}. Specifically, for step 1, the training of $E_\mathcal{X}$, $G_\mathcal{X}$, and $G_\mathcal{Y}$ are parallel. However, the training of step 2 is based on the results of step 1, and so as step 3 and 4. The batch size is set to 16 for each step. We initialize the network parameters using the Xavier initialization \cite{glorot2010understanding}. We use Adam optimizer \cite{kingma2014adam} with $\beta_1$=0.5 and $\beta_2$=0.999. The discriminators share the same learning rates as corresponding generators. There is no weight decay used in the training procedure, but the learning rates for individual steps are different, as listed in Table \ref{training_step_details}. There is no regularization terms used for the training.

\textbf{Dataset.} We adopt the training set of FF-Database (7148 parent and 8190 child faces) and 5000 high-quality child faces from FFHQ dataset \cite{karras2019style}. The FFHQ child faces are post-processed based on the same pipeline as FF-Database, e.g., 1) We use dlib to extract and align faces; 2) We label them with the same attributes as in FF-Database. All parent and child faces (including FF-Database and FFHQ) are used in the domain-specific training of domain $\mathcal{X}$ and $\mathcal{Y}$, respectively. The motivation to use FFHQ child faces is to enhance the generation quality of $G_\mathcal{Y}$.

The attributes used in the training for parent domain $\mathcal{X}$ are gender, moustache, glasses and expression, while they are age, gender, glasses and expression for children domain $\mathcal{Y}$. To simplify the computation of the classification losses, we define age and expression as binary attributes (i.e., young or senior and laugh or not laugh, respectively).

\textbf{Loss Function.} The loss functions used in each step are concluded in Table \ref{training_step_details}. The coefficients of different loss terms are shared for different steps, i.e., $\lambda_\mathcal{X}^1$, $\lambda_\mathcal{X}^2$, $\lambda_\mathcal{X}^3$, $\lambda_\mathcal{Z_X}^4$, $\lambda_\mathcal{Y}^1$, $\lambda_\mathcal{Y}^2$, $\lambda_\mathcal{Y}^3$, $\lambda_\mathcal{T}^1$, $\lambda_\mathcal{T}^2$, and $\lambda_\mathcal{T}^3$ are empirically set to 100, 10, 1, 0.1, 1, 1, 5, 0.8, 0.6, and 0.4, respectively.

\textbf{Time.} The ChildPredictor is trained on 8 NVIDIA Titan Xp GPUs (12 Gb memories for each). It is implemented by PyTorch 1.1.0 framework and Python 3.6. The training time of PGGAN $G_\mathcal{Y}$ is approximately 5 days. Considering the parallel training procedure (see Table \ref{training_step_details}), the remaining training time takes approximately 7 days.

\begin{table}[t]
\caption{The training details of the baseline CycleGAN. ``$G_{p \rightarrow c}$'' and ``$G_{c \rightarrow p}$'' represent the generators from the parent domain to the children domain and its reverse, respectively. ``$D_{p}$'' and ``$D_{c}$'' are the discriminators for parent and children domains, respectively. ``input Nc'' and ``output Nc'' are input and output images with N channels, respectively.}

\vspace{-3mm}

\label{baseline_details}
\begin{center}
\begin{tabular}{l|cc}
\hline
Item & Original CycleGAN & Baseline CycleGAN \\
\hline
$G_{p \rightarrow c}$ & input 3c, output 3c & input 6c, output 3c \\
$G_{c \rightarrow p}$ & input 3c, output 3c & input 3c, output 6c \\
$D_{p}$ & input 3c & input 6c \\
$D_{c}$ & input 3c & input 3c \\
Training data & random parent-child data & paired parents-child data \\
Parameters & the same as \cite{zhu2017unpaired} & the same as \cite{zhu2017unpaired} \\
\hline
\end{tabular}
\end{center}

\vspace{-4mm}

\end{table}

\subsection{Experiment Settings}

\subsubsection{Baselines}

To give a comprehensive and fair comparison, we do experiment on both domain transferring methods and state-of-the-art child face prediction DNA-Net. They are,
\begin{enumerate}\setlength{\itemsep}{-0.0cm}
\item I2I: DualGAN \cite{yi2017dualgan}, CycleGAN \cite{zhu2017unpaired}, UNIT \cite{liu2017unsupervised}, DRIT \cite{lee2018diverse}, MUNIT \cite{huang2018multimodal}, and DRIT++ \cite{lee2020drit++}, where DRIT, MUNIT, and DRIT++ predict multimodal by changing latent style codes;

\item DNA-Net \cite{gao2021dna}: It predicts multimodal by changing the linear factor in the random selection $S$.
\end{enumerate}

Since input and output are paired, we adjust the training scheme of all baselines by feeding parent-child pairs instead of randomly choosing samples from the whole training dataset for fairness. An example of I2I method is given in Table \ref{baseline_details}. Since DNA-Net is not open-source, we train it with the same attributes as ChildPredictor, where fathers and mothers are separately encoded, as shown in Figure \ref{com_arch} (b). In addition, to avoid ChildPredictor (using glasses, emotion, age and gender) and DNA-Net (using age and gender) using different face attributes in comparison, we fix the child external factors equal to ground truth for fairness.


\subsubsection{Evaluation}

We perform experiments on 170 validation parent-child pairs of the FF-Database. They are of the same image resolution as training data. Normally, the generated and ground truth faces are not strictly unaligned. To evaluate the generation quality, it is more reasonable to compare feature-level face similarity. To evaluate the generation diversity, we evaluate how large the differences are among different generated child faces from the same parent. For every family and every multimodal method, we predict 40 different child faces, which forms 40 groups of results. The experiments of subsections \emph{C-H} are performed on FF-Database validation set.

\textbf{Cosine distance (Cos. Dis.).} It measures feature-level cosine similarity between two faces. We use features from the second-last fully-connected layer of a Sphere20 network \cite{liu2017sphereface, Wang2018CosFace} pre-trained on CelebA dataset \cite{liu2015deep}. We compute pairwise cosine similarity for every family in every group and then compute the average of 40 groups. In order to demonstrate the effectiveness of cosine similarity, we randomly shuffle the 170 real parent-child pairs to obtain 6800 random parent-child pairs. Then, we compute the average of all the cosine similarity values. The average is 0.3204.

\textbf{Fréchet Inception Distance (FID) \cite{Heusel2017GANs}.} It measures the distance between two sets of images, i.e., the generated child faces and training data. The training data denotes child faces in the FF-Database training set. We use the features from the default ``pool3'' layer of the Inception-V3 \cite{Szegedy2016Rethinking} pre-trained on ImageNet \cite{deng2009imagenet}. We compute the FID for every group and then compute the average of 40 groups.


\textbf{Learned Perceptual Image Patch Similarity (LPIPS) \cite{Zhang2018The}.} It measures the diversity of the generated child faces. It is represented by the L1 distance between the features extracted from the AlexNet \cite{krizhevsky2012imagenet} pre-trained on ImageNet \cite{deng2009imagenet}. We compute the average of the pairwise distances among 40 groups of outputs for every family (i.e., 780 pairs if 40 outputs, $C_{40}^2 = 780$). Then, we compute the average across all 170 families. Note that, the color change is unwanted in child face prediction, although it facilitates the output diversity. For accurate evaluation, we conduct histogram equalization on all the generated images of all algorithms in comparisons. The histogram equalization operation removes the effect of color shift obviously. Thus, the evaluation is more fair in terms of the baselines (i.e., the style transfer methods often change the skin color, which is not real for this task).

\textbf{Human Perceptual Study.} We perform a human perceptual study to subjectively evaluate different methods. If a method obtains higher preference rates, it shows the method can produce higher-quality and more diverse faces. There are overall 14 human observers. The generated faces, input parents, and real child faces are presented to observers. The observers need to select one method that predicts faces closest to ground truth for each family. Finally, we count the preference rates.

\begin{figure*}[t]
\centering
\includegraphics[width=\linewidth]{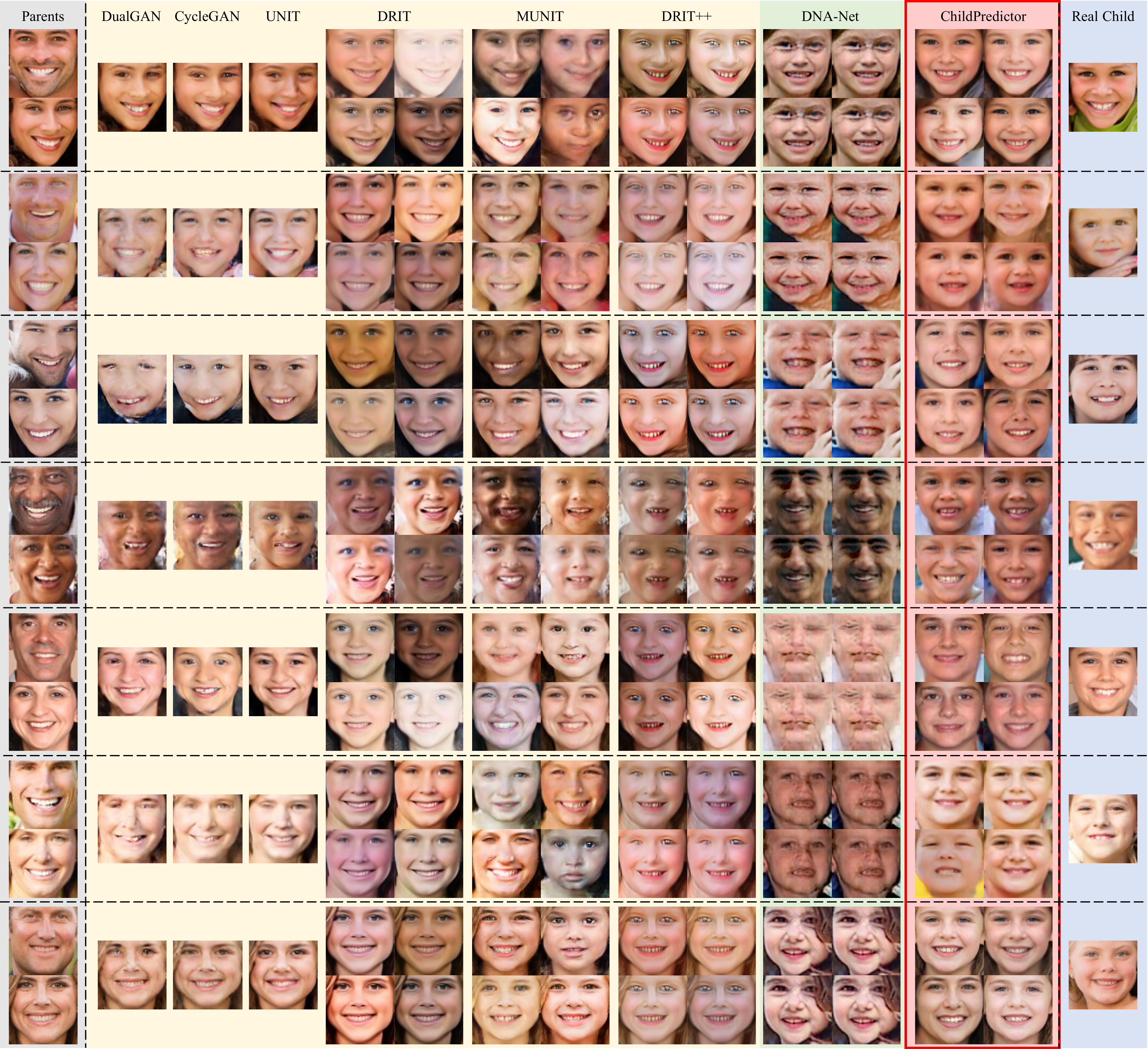}

\vspace{-3mm}

\caption{Illustration of some generated samples by ChildPredictor and baselines. The 1st column is input parents and the last column is real children. The 2nd-4th columns include parent-child pairs and single predicted children by different methods, respectively. The 5th-9th columns include diverse children generated by representative I2I methods (yellow background), DNA-Net (green background), and ChildPredictor (red background), respectively.}
\label{sota}

\vspace{-3mm}

\end{figure*}

\begin{table}[t]
\caption{Quantitative analysis of baselines and our ChildPredictor on Cosine distance, FID, and LPIPS metrics. The best performances are highlighted with the \textcolor{red}{red} color.}
\label{sota_metric}

\vspace{-3mm}

\begin{center}
\begin{tabular}{l|ccc}
\hline
Method & Cos. Dis. $\uparrow$ & FID $\downarrow$ & LPIPS $\uparrow$ \\
\hline
DualGAN & 0.3733 & 82.01 & / \\
CycleGAN & 0.3805 & 71.23 & / \\
UNIT & 0.3727 & 71.78 & / \\
\hline
DRIT & 0.3843 & 62.93 & 0.0041 \\
MUNIT & 0.3702 & 63.61 & 0.1669 \\
DRIT++ & 0.2013 & 79.92 & 0.0056 \\
\hline
DNA-Net & 0.3137 & 88.23 & 0.0087 \\
\hline
ChildPredictor (normal) & 0.4245 & 60.73 & 0.0063 \\
ChildPredictor (full) & \textbf{\textcolor{red}{0.4303}} & \textbf{\textcolor{red}{38.15}} & \textbf{\textcolor{red}{0.2757}} \\
\hline
\end{tabular}
\end{center}

\vspace{-5mm}

\end{table}

\begin{table}[t]
\caption{Subjective human perceptual study results of baselines and ChildPredictor on preference rates (PRs).}
\label{sota_subjective}

\vspace{-3mm}

\begin{center}
\begin{tabular}{l|cc}
\hline
Method & Similarity PR& Diversity PR  \\
\hline
DualGAN & 0.84\% & / \\
CycleGAN & 1.01\% & / \\
UNIT & 4.08\% & / \\
\hline
DRIT & 7.61\% & 5.04\% \\
MUNIT & 6.09\% & 9.83\% \\
DRIT++ & 1.21\% & 1.14\% \\
\hline
DNA-Net & 5.13\% & 1.05\% \\
\hline
ChildPredictor (full) & \textbf{\textcolor{red}{74.03\%}} & \textbf{\textcolor{red}{82.94\%}} \\
\hline
\end{tabular}

\vspace{-5mm}

\end{center}
\end{table}

\subsection{Experiment on Child Face Prediction Reality}

\subsubsection{Qualitative Analysis}

We illustrate some generated samples by different methods in Figure \ref{sota}. There are 4 samples for each multimodal method in the figure, where we sample different style codes for baselines to generate multiple faces, while we change genetic factors or variety factors for ChildPredictor.

Firstly, results from I2I methods (yellow background) are very similar to mothers. We claim it is an ``appearance collapse'' issue caused by the shared content space assumption. It promotes the networks simply copying the face structures from mothers; thereby the results are not similar enough compared with real children. Though we feed parent-child pairs to train baselines (see Figure \ref{baseline_details}), the disentanglement of content and style is not appropriate for this task. Secondly, DNA-Net assumes parents and children share the same content space like I2I methods but additionally uses a mapper to learn the relation between parent content codes and child content codes. After that, it performs an age-regression to the mapped child content codes to obtain a face. It is not like a natural biological process thus leading to artifacts in the generated faces.

ChildPredictor predicts very similar results with real children (e.g., face structure, color, and facial features). Also, the generated faces have less artifacts than baselines. We claim it is because the proposed disentangled learning is more accurate than the separation of content and style. Since external and variety factors are disentangled from genetic factors, the learning between parent and children domains of ChildPredictor is only on genetic factors. Compared with style codes, genetic factors are a special design for this task.

\subsubsection{Quantitative Analysis}

The quantitative results are concluded in Table \ref{sota_metric}. Since the I2I baselines do not adopt face attributes and FFHQ child data, we exclude them at the training stage for fairness, i.e., ``ChildPredictor (normal)''. Firstly, ``ChildPredictor (normal)'' obtains better cosine similarity and FID than the baselines. It demonstrates the architecture predicts the closest samples with ground truth with the best perceptual quality. Secondly, the full ChildPredictor enhances the results of ``ChildPredictor (normal)''. It is because of the use of attributes for the disentangled learning, which denotes the prediction-irrelevant information is disentangled from genetic factors. In addition, ChildPredictor obtains the highest LPIPS. It demonstrates that the design of predicting 4 genetic factors by $T$ and variety factors are helpful for producing diverse faces.

\subsubsection{Human Perceptual Study}

The preference rates (PRs) for each method in terms of both similarity and diversity are concluded in Table \ref{sota_subjective}. The ChildPredictor obtains clearly higher PRs than baselines, which demonstrates that it predicts perceptually more realistic and diverse faces, respectively.

\subsection{Ablation Study}

We conduct 9 experiments to evaluate several key components of ChildPredictor. The comparison results are included in the Table \ref{ablation_metric} and Figure \ref{ablation}. The analysis is as follows:

\begin{figure*}[t]
\centering
\includegraphics[width=\linewidth]{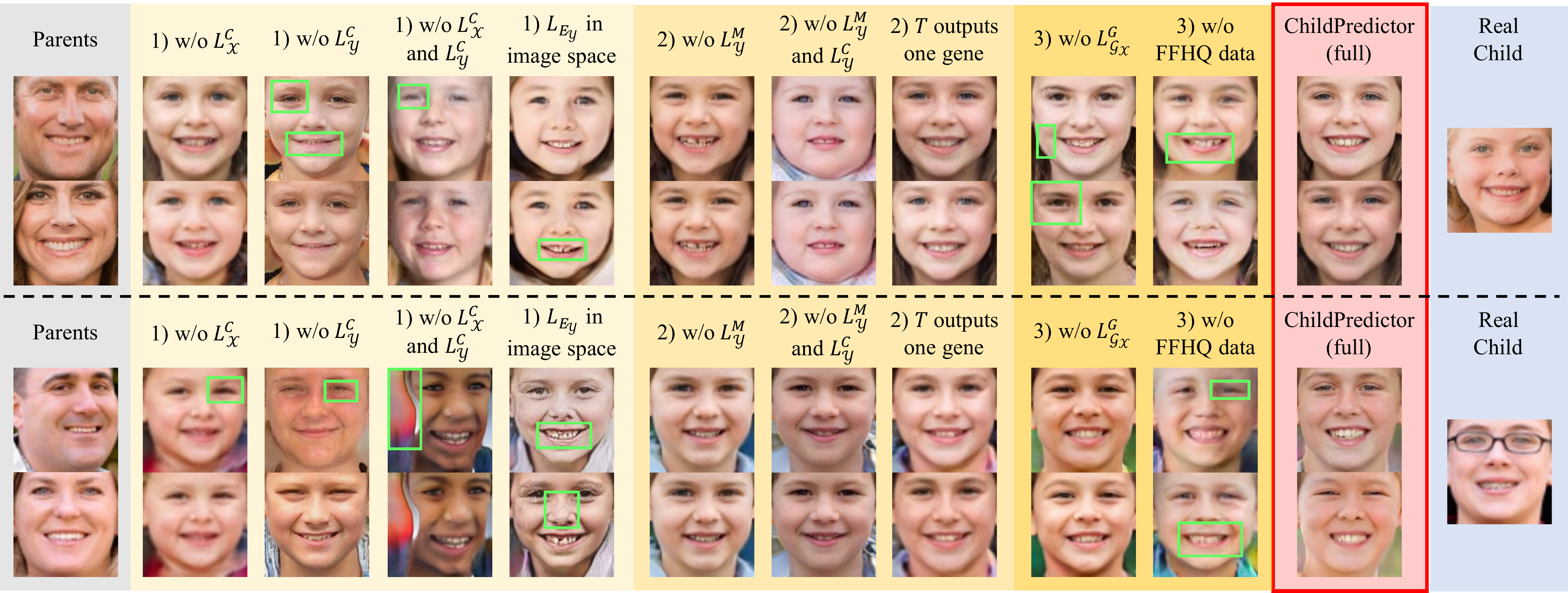}

\vspace{-3mm}

\caption{Illustration of the generated images under 9 different ablation study settings. There are 2 output images for each setting by only changing the variety factor. The left column is input parents (gray background), the middle part includes results of different settings  (yellow background), the right parts are full ChildPredictor's results and real child faces (red and blue background, respectively).}
\label{ablation}

\vspace{-4mm}

\end{figure*}

\begin{figure*}[t]
\centering
\includegraphics[width=\linewidth]{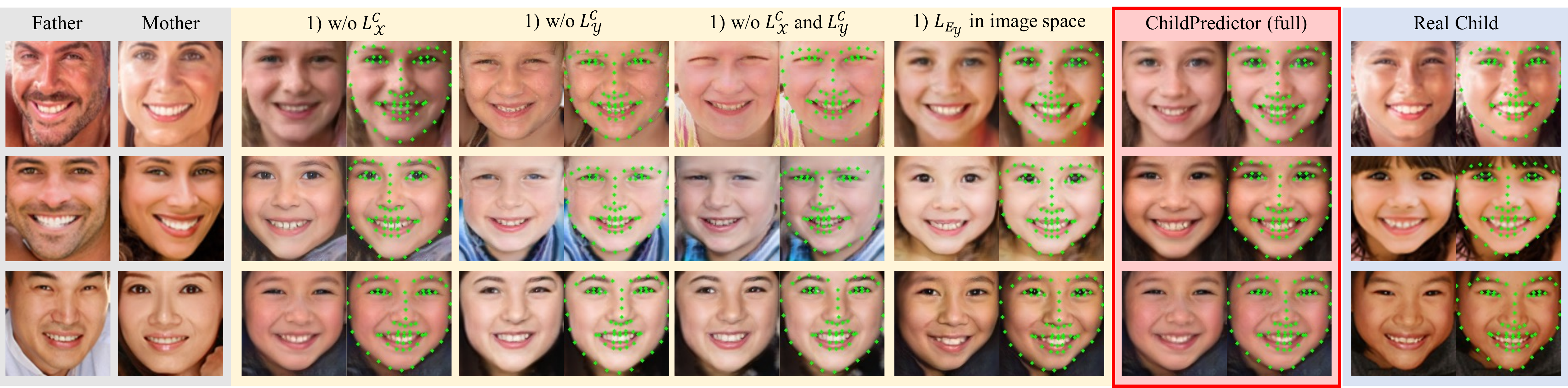}

\vspace{-3mm}

\caption{Illustration of more results on ChildPredictor with and without disentangled learning (i.e., ablation study setting 1)). The left two columns are inputs for the ChildPredictor. The 68 face landmarks (extracted by dlib) are illustrated alongside the generated child faces and real child faces.}
\label{more_disentangled_learning}

\vspace{-4mm}

\end{figure*}

\begin{table}[t]
\caption{Comparison of different ablation study settings of the proposed ChildPredictor.}
\label{ablation_metric}

\vspace{-5mm}

\begin{center}
\begin{tabular}{l|ccc}
\hline
Ablation study setting & Cos Dis. $\uparrow$ & FID $\downarrow$ & LPIPS $\uparrow$ \\
\hline
1) w/o $L_\mathcal{X}^C$ & 0.4167 & 43.76 & 0.2875 \\
1) w/o $L_\mathcal{Y}^C$ & 0.3665 & 39.71 & 0.2161 \\
1) w/o $L_\mathcal{X}^C$ and $L_\mathcal{Y}^C$ & 0.3572 & 39.89 & 0.2306 \\
1) $L_{E_\mathcal{Y}}$ in image space & 0.3592 & 39.39 & \textbf{\textcolor{red}{0.3511}} \\
\hline
2) w/o $L_\mathcal{Y}^M$ & 0.4247 & 54.51 & 0.2213 \\
2) w/o $L_\mathcal{Y}^M$ and $L_\mathcal{Y}^C$ & 0.4168 & 64.14 & 0.2213 \\
2) $T$ outputs one gene & 0.4161 & 52.57 & 0.0653 \\
\hline
3) w/o $L_{\mathcal{G_X}}^G$ & 0.4303 & 45.94 & 0.2236 \\
3) w/o FFHQ data & 0.4296 & 69.11 & 0.1482 \\
\hline
ChildPredictor (full) & \textbf{\textcolor{red}{0.4303}} & \textbf{\textcolor{red}{38.15}} & 0.2757 \\
\hline
\end{tabular}

\vspace{-5mm}

\end{center}
\end{table}

\subsubsection{Disentangled Learning Ability}

We exclude the classification losses $L_\mathcal{X}^C$ or $L_\mathcal{Y}^C$ or both to train ChildPredictor without external factors. Also, we optimize $E_\mathcal{Y}$ in image space rather than on genetic factors (i.e., $L_{E_\mathcal{Y}}$ performs on images), leading to false disentangled learning in children domain. All the settings lead to obvious decreases of metrics (e.g., more than 0.06 decrease of Cos. Dis for ``w/o $L_\mathcal{Y}^C$''). The outputs are not realistic, even contain visual artifacts, e.g., the faces, eyes and mouths marked by \textcolor{green}{green} rectangles of Figure \ref{ablation} 1).

\subsubsection{Generation Diversity}

We exclude the mode-seeking loss $L_\mathcal{Y}^M$ or let mapping function $T$ produce one genetic factor instead of 4 factors. The LPIPS metric of these settings decreases obviously due to no effect of variety factors or no diverse output genetic factors. Therefore, those different generated faces are almost the same, as shown in Figure \ref{ablation} 2).

\subsubsection{Other Terms}

We exclude the auxiliary loss $L_{\mathcal{G_X}}^G$ or additional FFHQ data when training $G_\mathcal{Y}$. All the metrics decrease since $L_{\mathcal{G_X}}^G$ and FFHQ data contribute to high-quality image generation. The predicted faces are blurry or ghosted if excluding them, as the specific regions marked by \textcolor{green}{green} rectangles of Figure \ref{ablation} 3).

In conclusion, all the network components, loss functions, and disentangled learning method are significant for ChildPredictor to generate realistic and diverse child faces.

\subsection{Experiment on Disentangled Learning}

To further demonstrate the effectiveness of disentangled learning, we show more predicted faces with and without disentangled learning (w/o $L_\mathcal{X}^C$, w/o $L_\mathcal{Y}^C$, w/o $L_\mathcal{X}^C$ and $L_\mathcal{Y}^C$, and $L_{E_\mathcal{Y}}$ in image space; please see ablation study setting 1)). The results are illustrated in Figure \ref{more_disentangled_learning}, where 68 landmarks are illustrated for every face. The results from full ChildPredictor are more similar to real children.

\begin{figure*}[t]
\centering
\includegraphics[width=\linewidth]{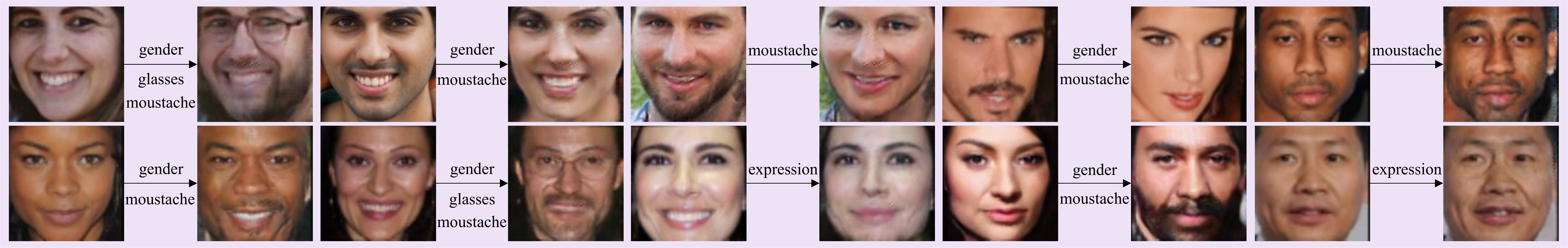}

\vspace{-3mm}

\caption{Disentangled learning in parent domain $\mathcal{X}$. The left is input and the right is the generated face. The texts around arrows denote changed attributes.}

\vspace{-3mm}

\label{disentanglex}
\end{figure*}

\vspace{-3mm}

\begin{figure*}[t]
\centering
\includegraphics[width=\linewidth]{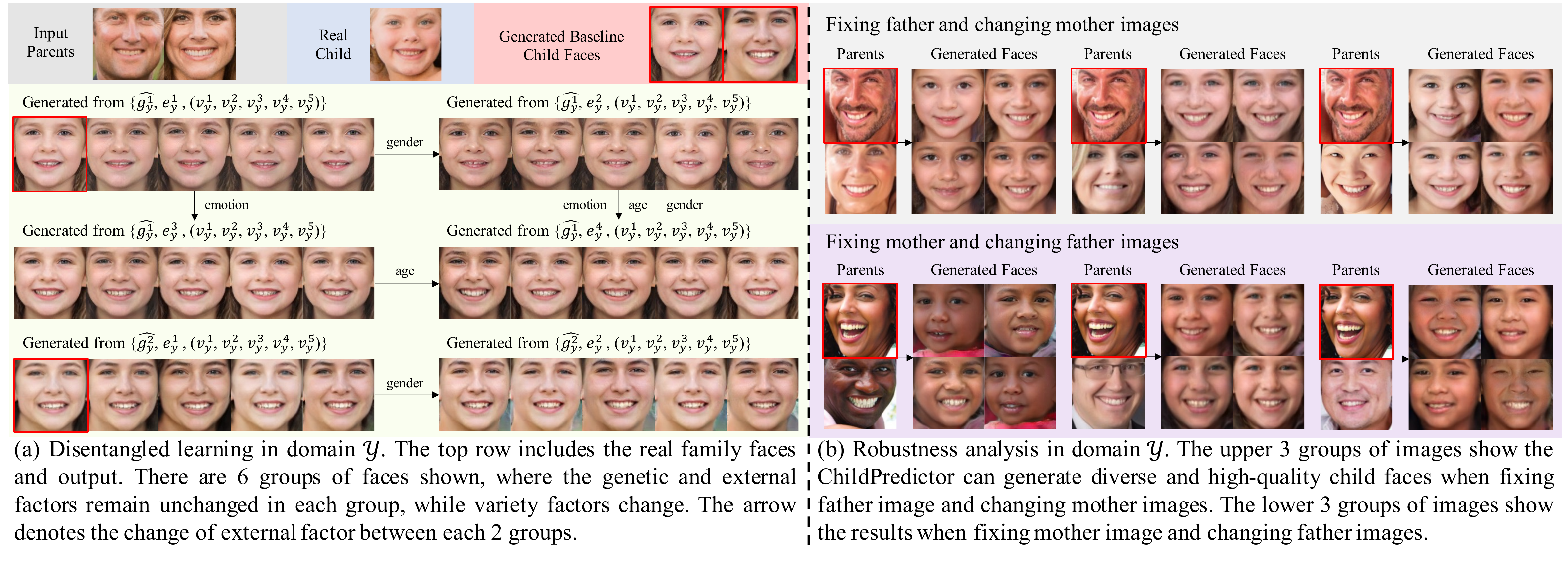}
\vspace{-7mm}

\caption{Illustration of (a) Disentangled learning in children domain $\mathcal{Y}$: generated faces of the same parents from different genetic, external, and variety factors; (b) Robustness analysis in children domain $\mathcal{Y}$: fixing one of input parents and changing the other.}
\label{disentangley}

\vspace{-2mm}

\end{figure*}

\begin{figure*}[t]
\centering
\includegraphics[width=\linewidth]{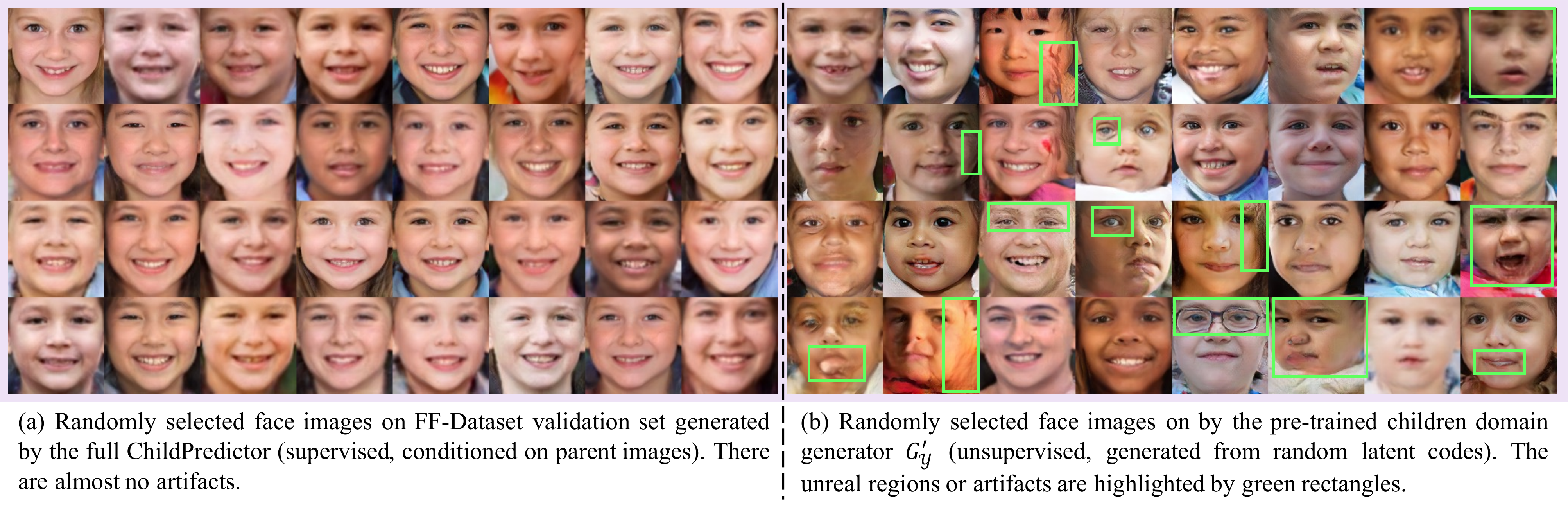}

\vspace{-4mm}

\caption{Illustration of child face prediction results by supervised ChildPredictor and the unsupervised pre-trained children domain generator $G_\mathcal{Y}^{'}$. There are 32 images shown for each setting, which are randomly sampled from 6800 predicted face images.}

\vspace{-3mm}

\label{supervised_learning}
\end{figure*}

\subsection{Disentangled Learning and Robustness Analysis}

The disentangled learning is the basic of ChildPredictor since it assists to extract accurate genetic factors. For parent domain, we walk the latent code of $g_x$ or $e_x$ and fix the other, as shown in Figure \ref{disentanglex}. Obviously, changing genetic factors $g_x$ will not influence attributes, and modifying attributes $e_x$ will not change identity. Therefore, the parent genetic factors can well represent prediction-relevant information. For children domain, we illustrate generated faces from one input parent, e.g., from different genetic factors $\widehat{g_y^j}$ predicted by mapping function or different external factors $e_y^j$ and variety factors $v_y^j$, as shown in Figure \ref{disentangley} (a). The $e_y^j$ only changes specific attributes, while the $v_y^j$ only influences individual properties when other factors are fixed. We also show generated faces by changing one of input parents, as shown in Figure \ref{disentangley} (b). The model is robust to different inputs, which proves that it does not fall into the appearance collapse.

\begin{table}[t]
\caption{Comparison of the FID for the full ChildPredictor and the pre-trained children domain generator $G_\mathcal{Y}^{'}$ on validation set.}
\label{com_supervised_learning}

\vspace{-3mm}

\begin{center}
\begin{tabular}{l|c}
\hline
Method & FID $\downarrow$ \\
\hline
Full ChildPredictor (supervised learning) & \textbf{\textcolor{red}{50.77}} \\
Pre-trained $G_\mathcal{Y}^{'}$ (unsupervised learning) & 53.89 \\
\hline
\end{tabular}

\vspace{-4mm}

\end{center}
\end{table}

\subsection{Supervised Learning Analysis}

To demonstrate the information from parents improves the child prediction quality, we compare results from the full ChildPredictor framework with randomly generated samples from the pre-trained PGGAN ($G_\mathcal{Y}^{'}$). The parent-child pairs are utilized for supervised learning for our framework. Note that $G_\mathcal{Y}^{'}$ is pre-trained in an unsupervised manner (following PGGAN training as Equation \ref{335}); therefore, the weights are not the same as $G_\mathcal{Y}$ in the ChildPredictor framework. There are overall 6800 generated face images on validation set by ChildPredictor and we also randomly generate 6800 samples by pre-trained $G_\mathcal{Y}$. To compare the image generation quality of supervised learning and unsupervised learning, we compute the FID for them on validation set and the results are concluded in Table \ref{com_supervised_learning}. It shows that results from the full ChildPredictor are more similar to real child faces in the validation set than the pre-trained PGGAN $G_\mathcal{Y}^{'}$.

In addition, we illustrate the generated results of the full ChildPredictor and $G_\mathcal{Y}^{'}$ in Figure \ref{supervised_learning}. It is obvious that the children domain generator can predict faces with better quality conditioned on parent images. The supervised learning provides a more fixed latent space for children domain generator $G_\mathcal{Y}$; therefore, it can produce more reasonable faces.

\begin{figure*}[t]
\centering
\includegraphics[width=\linewidth]{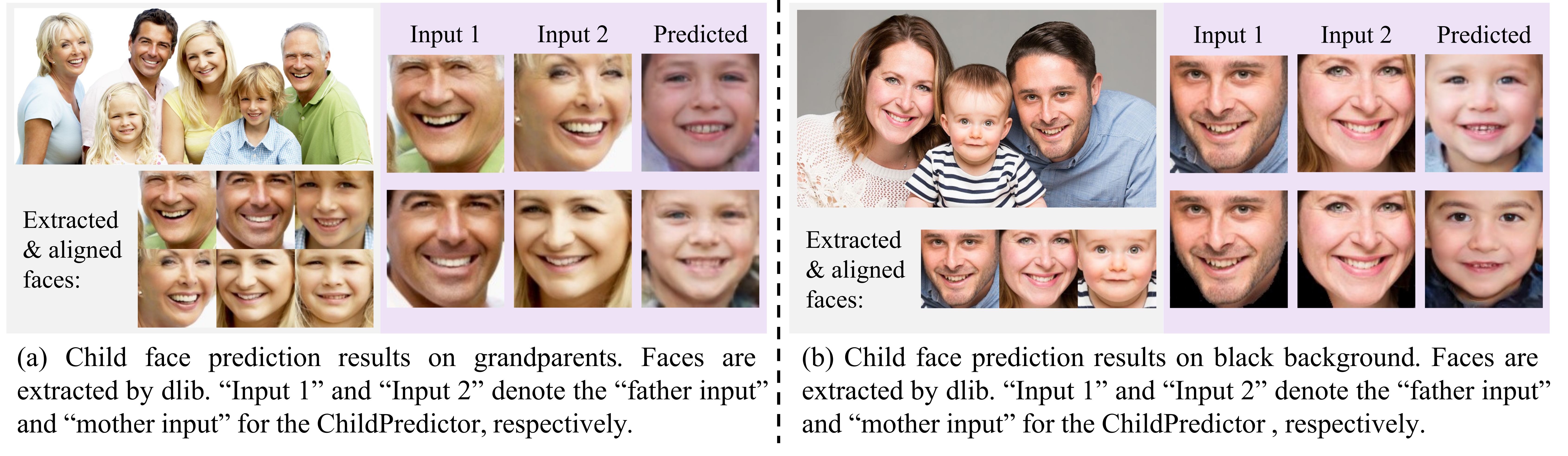}

\vspace{-3mm}

\caption{Illustration of child face prediction results on grandparents and black background, respectively. The original images are also shown for reference.}

\vspace{-1mm}

\label{other_circumstance}
\end{figure*}

\begin{figure*}[t]
\centering
\includegraphics[width=17cm, height=9.5cm]{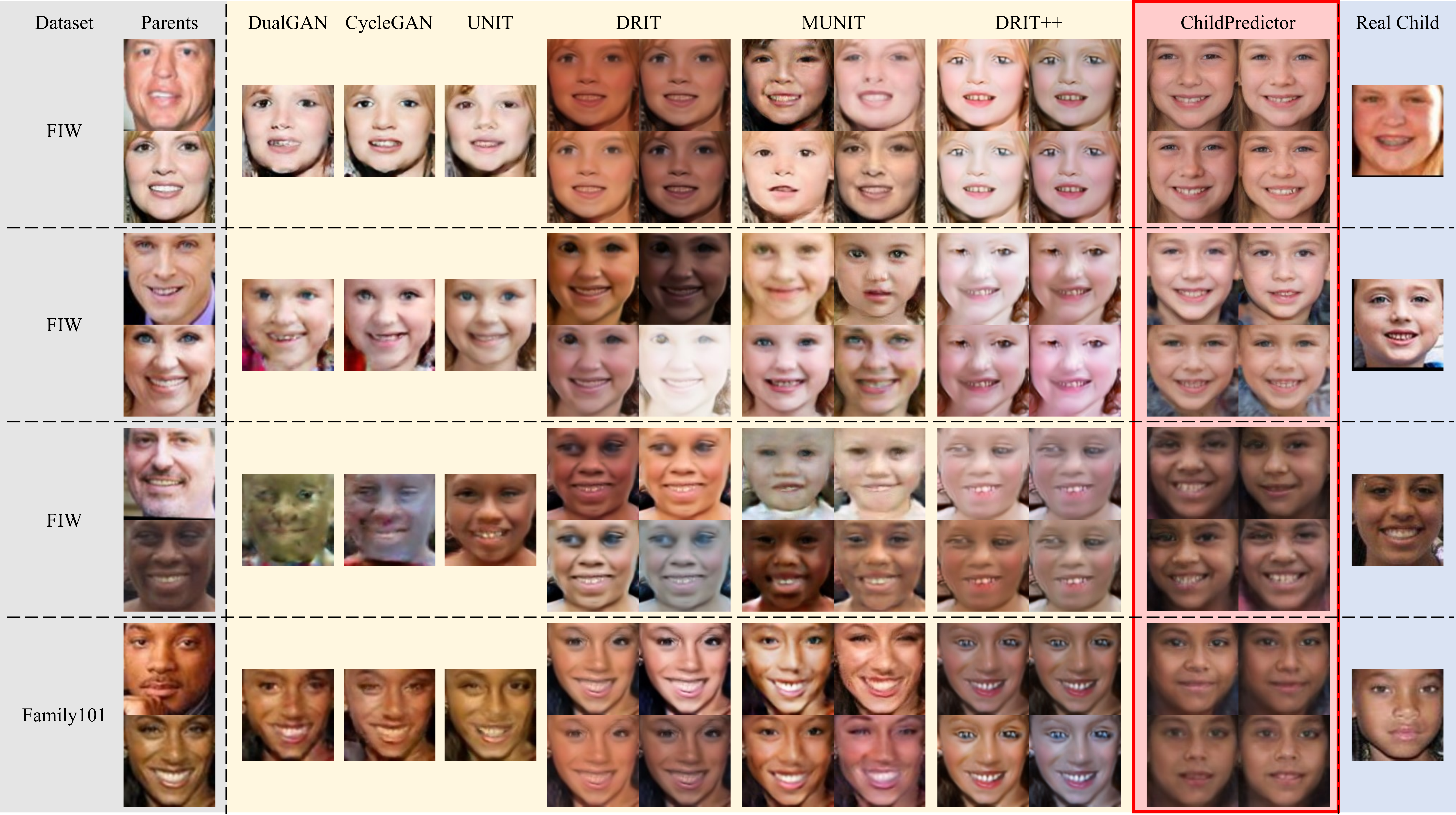}
\vspace{-3mm}

\caption{Illustration of some generated samples by ChildPredictor (red background) and I2I baselines (yellow background). Input parents and real children are shown in the left and right, respectively. The samples are selected from the processed FIW dataset and Family101 dataset, respectively.}
\label{other_dataset}

\vspace{-3mm}

\end{figure*}

\subsection{Experiment on Other Circumstances}

We consider two other circumstances: 1) The ``parents'' are replaced by ``grandparents'' and 2) The background is darkened. The predicted child faces are illustrated in Figure \ref{other_circumstance}, where results from the normal setting are also shown for comparisons. In case 1), the ChildPredictor can still output child face given grandparent faces. It is because the mapping function can still map the grandparents' genetic factors to child genetic factors in the learned space, while the children domain generator transforms the predicted child genetic factors to child faces. In case 2), the ChildPredictor can still predict a face similar with ground truth when changing the background color  (e.g., the background color is darkened in Figure \ref{other_circumstance} (b)). However, the two predicted child faces under normal lighting condition and dark background are not very similar. It is because different background colors are not modeled during the training. In future work, we will consider it and make the ChildPredictor more robust.

\subsection{Experiment on Other Datasets}

To further evaluate the proposed ChildPredictor, we include two more datasets: Families in the Wild (FIW)\footnote{https://web.northeastern.edu/smilelab/fiw/} \cite{robinson2018visual} and Family101\footnote{http://chenlab.ece.cornell.edu/projects/KinshipClassification/index.html} \cite{fang2013kinship}. However, the original images in the datasets are not processed with the same pipeline as FF-Database, or have different image resolutions and formats (e.g., grayscale format). To minimize the gap, we apply the same pre-processing procedures to the images in the two datasets. In addition, we manually exclude some grandparent-parent pairs and some profile face images. The pre-processing procedures result in 50 validation pairs from the FIW dataset (extracted from ``F0001'' to ``F0200'' of FIW training images) and 17 validation pairs from the Family101 dataset, respectively. They are of 128$\times$128 resolution and not overlapped with FF-Database images. The processed images will be publicly available.

Since the validation sets are relatively small, we only adopt the cosine distance and LPIPS as quantitative metrics because they evaluate the pairwise image quality. The baselines and ChildPredictor trained on the FF-Database are used in the experiment. Note that the DNA-Net cannot well generalize to FIW and Family101 images (e.g., predicted faces are extremely ambiguous) so we do not include it in the experiment. The quantitative results are concluded in Table \ref{other_dataset_metric}. The proposed ChildPredictor achieves better performances than other methods according to face prediction similarity and diversity on both datasets. The results are consistent with the conclusion on the FF-Database validation set.

Some samples are illustrated in Figure \ref{other_dataset}. Compared with other methods, ChildPredictor predicts more similar faces to real children, more diverse faces, and higher-quality faces. For instance, I2I baselines (please see the yellow background in Figure \ref{other_dataset}) still produce faces with very similar shapes to input mothers, i.e., appearance collapse. However, ChildPredictor does not have this issue since the mapping is performed in the genetic domain, while the face-prediction-irrelevant factors are disentangled. The visual results also demonstrate that the proposed ChildPredictor has better generalization ability since it is not trained on these two datasets.

\begin{table}[t]
\caption{Quantitative analysis of baselines and our ChildPredictor on Cosine distance and LPIPS metrics. The results on FIW and Family101 are separately represented.}
\label{other_dataset_metric}

\vspace{-3mm}

\begin{center}
\begin{tabular}{l|cc|cc}
\hline
\multirow{2}{*}{Method} & \multicolumn{2}{c|}{FIW} & \multicolumn{2}{c}{Family101} \cr & Cos. Dis. $\uparrow$ & LPIPS $\uparrow$ & Cos. Dis. $\uparrow$ & LPIPS $\uparrow$ \\
\hline
DualGAN & 0.3533 & / & 0.3612 & / \\
CycleGAN & 0.3649 & / & 0.3771 & / \\
UNIT & 0.3793 & / & 0.3878 & / \\
\hline
DRIT & 0.3666 & 0.0047 & 0.3531 & 0.0049 \\
MUNIT & 0.3658 & 0.1727 & 0.3665 & 0.1715 \\
DRIT++ & 0.1907 & 0.0065 & 0.1648 & 0.0069 \\
\hline
ChildPredictor & \textbf{\textcolor{red}{0.4259}} & \textbf{\textcolor{red}{0.2247}} & \textbf{\textcolor{red}{0.3914}} & \textbf{\textcolor{red}{0.2323}} \\
\hline
\end{tabular}
\end{center}

\vspace{-3mm}

\end{table}

\subsection{Experiment on Famous Families}

\begin{figure}[t]
\centering
\includegraphics[width=\linewidth]{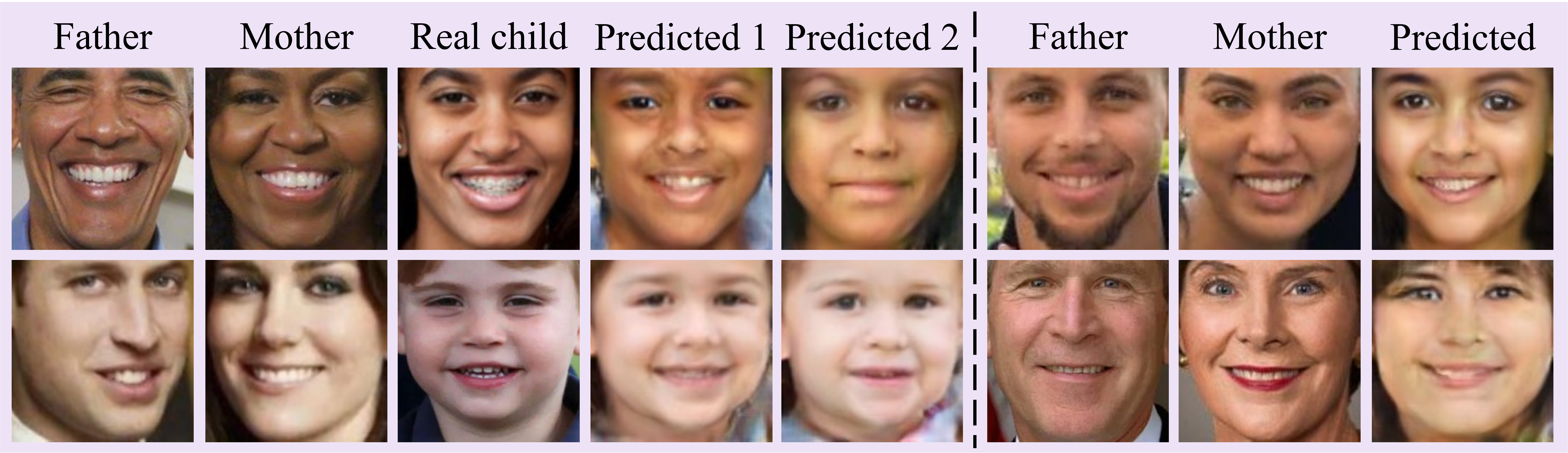}

\vspace{-1mm}

\caption{Illustration of child face prediction results on 4 famous family images downloaded from the Internet.}

\vspace{-1mm}

\label{famous_family}
\end{figure}

We download 4 famous family images from the Internet for testing the ChildPredictor. The same pre-processing pipeline is applied to these images. The predicted face images are illustrated in Figure \ref{famous_family}. The ChildPredictor framework can predict reasonable child faces for these real-world cases.

\subsection{Limitation of ChildPredictor Framework}

\begin{figure}[t]
\centering
\includegraphics[width=\linewidth]{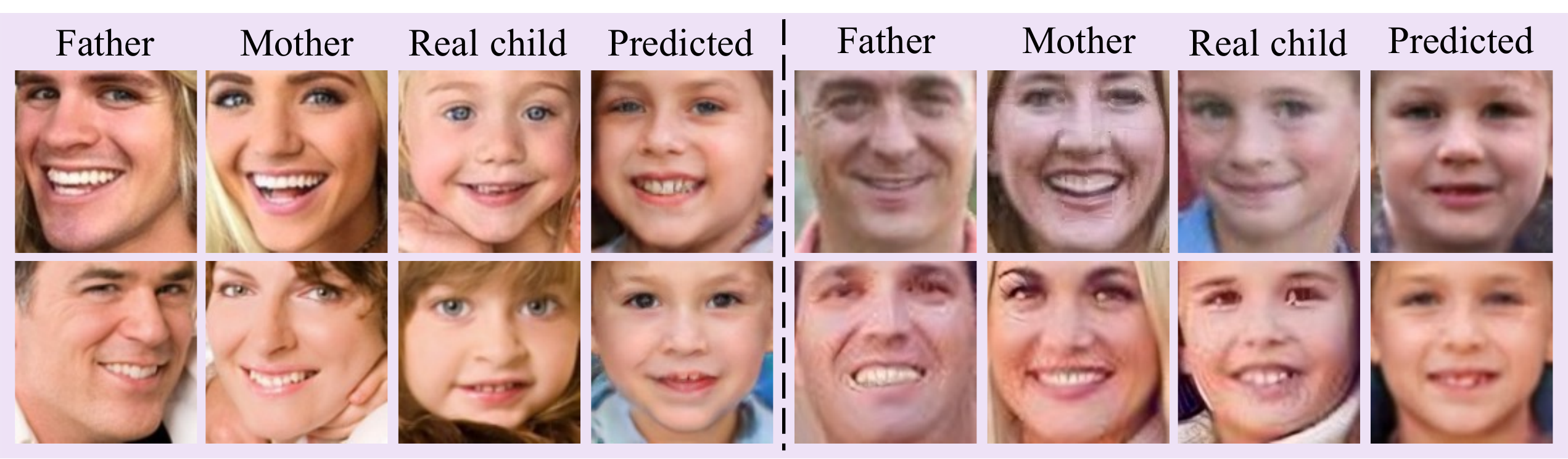}

\vspace{-1mm}

\caption{Illustration of common failure predicted faces.}
\label{fail}

\vspace{-1mm}

\end{figure}

For many situations, ChildPredictor can generate high-quality child faces. However, there are some common failure cases shown in Figure \ref{fail} including: side face input (left two samples) and low-quality input (right two samples). It may be because there are little profile or low-quality faces in FF-Database. In the future, we will enhance the ChildPredictor to be more robust to such input images.

\section{Broader Impact}

The success of ChildPredictor depends on the proposed large-scale FF-database. Although the faces in the dataset are anonymous, they may be sensitive to privacy issues. \emph{We are strongly aware that privacy protection is a significant issue in the community.} To largely protect privacy: 1) We plan not to disclose original face images of the FF-Database; however, \emph{we will alternatively release the features extracted by state-of-the-art face recognition networks for future study}. 2) In terms of application, we encode faces by an irreversible process and delete the background data; 3) In terms of algorithm, we can adopt privacy-preserving generative models \cite{xie2018differentially, lin2021privacy, chen2020gs, xiao2020adversarial}. These methods promote discoveries that may be hindered by data-protection barriers and maintain the reproducibility of the algorithm; 4) In terms of generated data, we can learn the characteristics of generated samples to identify them \cite{wang2020cnn}. They are orthogonal work and we will not include them in this paper. In the future, we will use them to enhance the privacy protection of ChildPredictor.

The ChildPredictor can synthesize realistic child faces from parents. It helps to solve many social issues, such as missing child identification and criminal investigations. However, to avoid improper use, we are very cautious about the pros and cons of ChildPredictor. We will strictly control the use of ChildPredictor in the aforementioned social applications.

\section{Conclusion}

In this paper, we presented a ChildPredictor framework to automatically predict diverse child faces from parents. In order to simulate a biological process, we formulate it as a genetic factor mapping problem. We learn this mapping from parents to children in the latent space. We adopt the encoder-generator architecture to connect the image spaces and latent spaces of the parent and children domains. To extract precise genetic factors, we exclude external factors (facial attributes) and variety factors (individual properties) based on disentangled learning. For the parent domain, it is achieved by enforcing classification loss on generated faces. For the children domain, it is implemented by regularizing the latent space by a GAN encoder. We collected a large-scale Family Face Database (FF-Database) to train the ChildPredictor. It includes 16046 faces (7148 parent faces and 8190 child faces) with labeled facial attributes. Finally, we validated the ChildPredictor with several state-of-the-art methods by both quantitative analysis and human perceptual study on the FF-Database. Experiment results demonstrate that our ChildPredictor can predict higher-quality, more diverse, and more realistic child faces than state-of-the-art methods.

\section*{Acknowledgment}

The authors would like to thank Kangcheng Liu, Qinbin Li, Zhanghan Ke, and Yurou Zhou for their reviews, and Su Wang for labeling a part of facial attributes in the proposed FF-Database. The authors would also like to thank the anonymous reviewers and editors for their helpful comments.

\ifCLASSOPTIONcaptionsoff
  \newpage
\fi



%

{
\bibliographystyle{IEEEtran}
\bibliography{IEEEabrv,mybibfile}
}

%
%

%

\vspace{-11mm}

\begin{IEEEbiography}[{\includegraphics[width=1in,height=1.25in,clip,keepaspectratio]{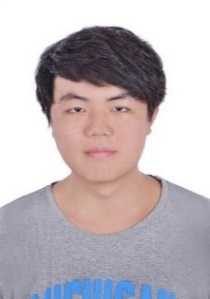}}]{Yuzhi Zhao}

(S’19) received the B.Eng. Degree in electronic information from Huazhong University of Science and Technology, Wuhan, China, in 2018. He is currently pursuing the Ph.D. degree with the Department of Electronic Engineering, City University of Hong Kong. His research interests include image processing and deep learning, particularly generative models, image \& video enhancement, and low-level \& physic-based computer vision.

\end{IEEEbiography}

\vspace{-6mm}

\begin{IEEEbiography}[{\includegraphics[width=1in,height=1.25in,clip,keepaspectratio]{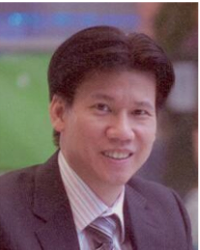}}]{Lai-Man Po}

(M’92–SM’09) received the B.S. and Ph.D. degrees in electronic engineering from the City University of Hong Kong, Hong Kong, in 1988 and 1991, respectively. He has been with the Department of Electronic Engineering, City University of Hong Kong, since 1991, where he is currently an Associate Professor of Department of Electrical Engineering. He has authored over 150 technical journal and conference papers. His research interests include image and video coding with an emphasis deep learning based computer vision algorithms.

Dr. Po is a member of the Technical Committee on Multimedia Systems and Applications and the IEEE Circuits and Systems Society. He was the Chairman of the IEEE Signal Processing Hong Kong Chapter in 2012 and 2013. He was an Associate Editor of HKIE Transactions in 2011 to 2013. He also served on the Organizing Committee, of the IEEE International Conference on Acoustics, Speech and Signal Processing in 2003, and the IEEE International Conference on Image Processing in 2010.

\end{IEEEbiography}

\begin{IEEEbiography}[{\includegraphics[width=1in,height=1.25in,clip,keepaspectratio]{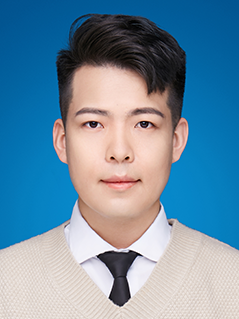}}]{Xuehui Wang}

(S’21) received the B.S. Degree from Shandong University, China, and the Master degree in the School of Computer Science from Sun Yat-sen Uninversity, China, in 2018 and 2021, respectively. He is currently pursuing the PhD degree at Artificial Intelligence Institute, Shanghai Jiao Tong University, China. His research interests include computer vision (super resolution, instance segmentation), deep learning.

\end{IEEEbiography}

\begin{IEEEbiography}[{\includegraphics[width=1in,height=1.25in,clip,keepaspectratio]{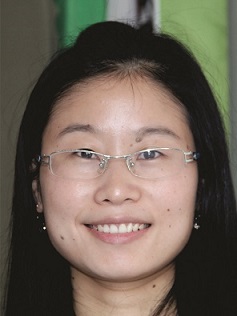}}]{Qiong Yan}

received her Ph.D. degree in computer science and engineering from Chinese University of Hong Kong in 2013 and the Bachelor’s degree in computer science and technology from University of Science and Technology of China in 2009. She is now a research director in SenseTime, leading a group on computational imaging related research and production. Her research focuses on low-level vision tasks, such as image/video restoration and enhancement, image editing and generation.

\end{IEEEbiography}

\begin{IEEEbiography}[{\includegraphics[width=1in,height=1.7in,clip,keepaspectratio]{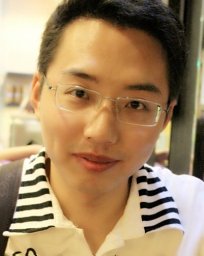}}]{Wei Shen}

is a tenure-track Associate Professor at the Artificial Intelligence Institute, Shanghai Jiao Tong University, since October 2020. Before that, he was an Assistant Research Professor at the Department of Computer Science, Johns Hopkins University. His research interests lie in the fields of computer vision, machine learning, deep learning, and medical image analysis. He serves as an Associate Editor for Neurocomputing and an Area Chair for CVPR 2022.

\end{IEEEbiography}

\begin{IEEEbiography}[{\includegraphics[width=1in,height=1.7in,clip,keepaspectratio]{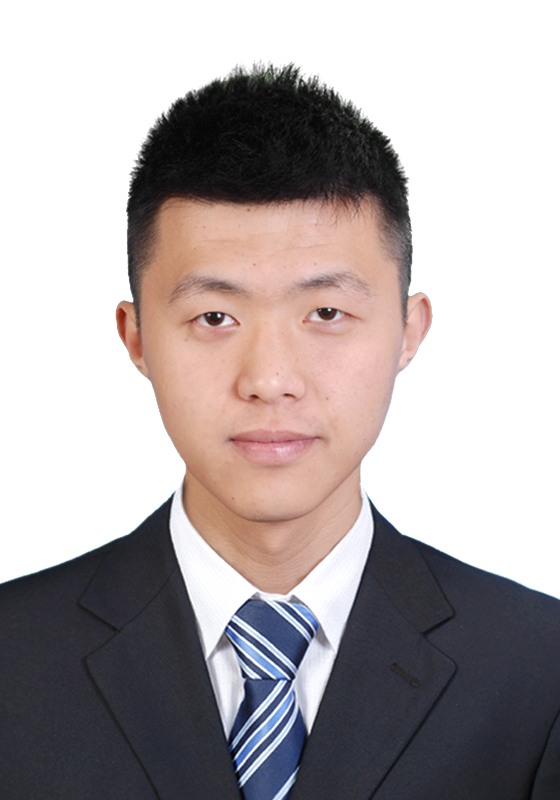}}]{Yujia Zhang}

received the B.E. degree in electrical engineering and automation in Huazhong University of Science and Technology in 2015, and the M.S. degree in electrical engineering in South China University of Technology, China, in 2018. He is currently pursuing the Ph. D. degree in City University of Hong Kong. His current research interests include computer vision, video understanding.

\end{IEEEbiography}

\begin{IEEEbiography}[{\includegraphics[width=1in,height=1.25in,clip,keepaspectratio]{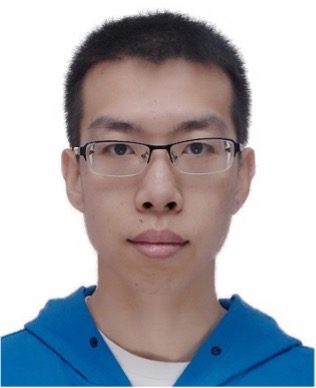}}]{Wei Liu}

received his B.S. and PhD degrees from Harbin Institute of Technology, Harbin, China, in 2016 and 2020, respectively. He was a visiting student in the Ohio State University for two years. He used to be an intern at SenseTime and currently works as an algorithm engineer at ByteDance. His research interests include image generation, domain adaptation, semantic segmentation and low-level computer vision. Dr. Liu serves as a Peer Reviewer for IEEE Transactions on Image Processing, ISPRS Journal of Photogrammetry and Remote Sensing, IEEE Transactions on Geoscience and Remote Sensing, etc.

\end{IEEEbiography}

\begin{IEEEbiography}[{\includegraphics[width=1in,height=1.7in,clip,keepaspectratio]{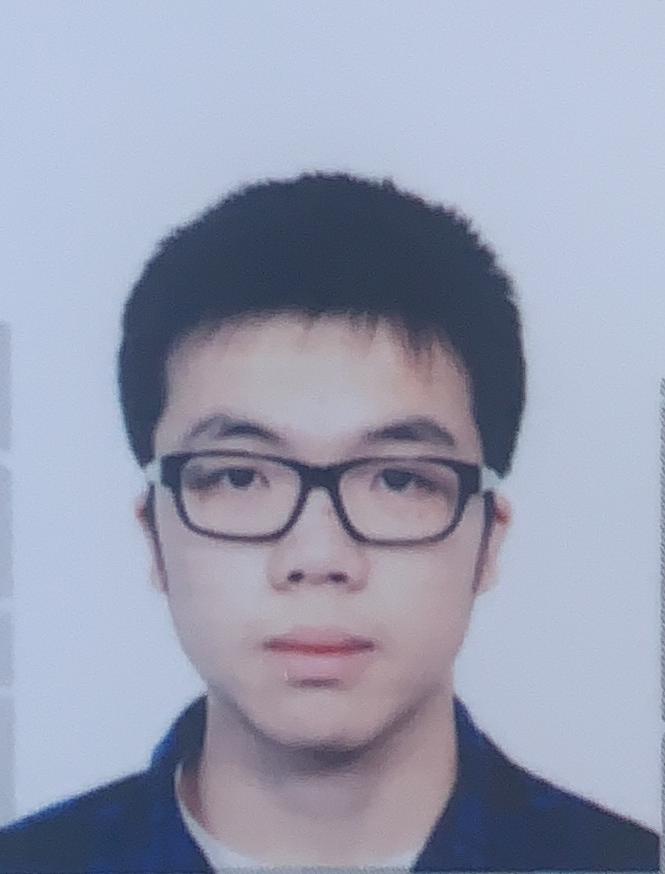}}]{Chun-Kit Wong}

is pursuing the B.Eng. Degree of Information Engineering in City University of Hong Kong. His research interests include deep learning and computer vision. 

\end{IEEEbiography}

\begin{IEEEbiography}[{\includegraphics[width=1in,height=1.7in,clip,keepaspectratio]{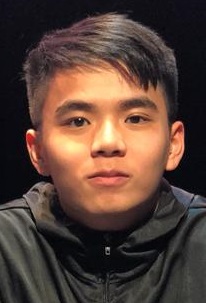}}]{Chiu-Sing Pang}

is pursuing the B.Eng. Degree of Engineering in Information Engineering in City University of Hong Kong. His research interests include deep learning and computer vision. 

\end{IEEEbiography}

\begin{IEEEbiography}[{\includegraphics[width=1in,height=1.7in,clip,keepaspectratio]{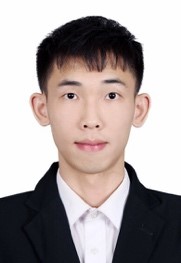}}]{Weifeng Ou}

received his B.Eng. degree in Telecommunication Engineering from Guangdong University of Technology in 2013, his M.Eng. degree in Signal \& Information Processing from South China University of Technology in 2016, and his Ph.D. degree in the Department of Electrical Engineering from City University of Hong Kong in 2021. He was with Huawei as an R \& D engineer from 2016 to 2018. He is currently working in Sensetime. His research interests include biometrics and deep learning. 

\end{IEEEbiography}

\begin{IEEEbiography}[{\includegraphics[width=1in,height=1.25in,clip,keepaspectratio]{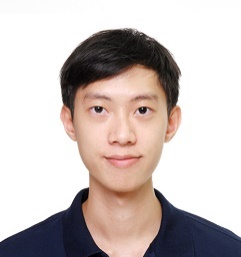}}]{Wing-Yin Yu}

received the B.Eng. degree in Information Engineering from City University of Hong Kong, in 2019. He is currently pursuing the Ph.D. degree at Department of Electrical Engineering at City University of Hong Kong. His research interests are deep learning and computer vision.

\end{IEEEbiography}

\begin{IEEEbiography}[{\includegraphics[width=1in,height=1.25in,clip,keepaspectratio]{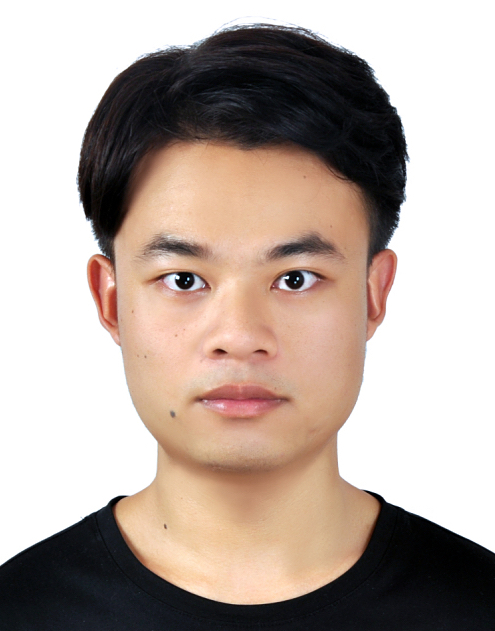}}]{Buhua Liu}

received the B.Eng. degree in School of Electronic Information and Communications from Huazhong University of Science and Technology, Wuhan, China, in 2019. He is currently pursuing the Ph.D. degree with the Department of Computer Science, Hong Kong Baptist University. His research interests lie in the fields of AI security, privacy and computer vision, particularly in adversarial learning, federated learning.

\end{IEEEbiography}

\end{document}